\documentclass[10pt,twocolumn,letterpaper]{article}

\usepackage{iccv}
\usepackage{times}
\usepackage{epsfig}
\usepackage{graphicx}
\usepackage{amsmath}
\usepackage{amssymb}

\usepackage{stmaryrd} 
\usepackage{multirow} 
\usepackage{ctable}
\usepackage{array}
\usepackage{subcaption}
\usepackage{enumitem} 
\usepackage[symbol]{footmisc}

\usepackage{ulem}
\usepackage{color}
\definecolor{crimson}{rgb}{0.86, 0.08, 0.24}
\definecolor{gray}{rgb}{0.5,0.5,0.5}
\definecolor{green}{rgb}{0, 0.4, 0}
\definecolor{orange}{rgb}{1, 0.5, 0}
\definecolor{mahogany}{rgb}{0.75, 0.25, 0.0}
\definecolor{purple}{rgb}{0.6, 0, 0.6}
\definecolor{darkgreen}{rgb}{0, 0.4, 0}
\definecolor{frenchblue}{rgb}{0.0, 0.45, 0.73}
\definecolor{red}{rgb}{1,0,0}
\definecolor{yellow}{rgb}{1,1,0}
\definecolor{magenta}{rgb}{1,0,1}
\definecolor{pink}{rgb}{1,0.412,0.706}

 \def \submission {}
\ifx \submission \undefined
    \newcommand{\johnson}[1]{{\color{frenchblue}{#1}}}
	
	\newcommand{\charles}[1]{{\color{orange}{#1}}}
	
	\newcommand{\hubert}[1]{{\color{pink}{#1}}}

	\newcommand{\todo}[1]{{\color{purple}{#1}}}
	
\else
    \newcommand{\johnson}[1]{{#1}}
	
	\newcommand{\charles}[1]{{#1}}
	
	\newcommand{\hubert}[1]{{#1}}

	\newcommand{\todo}[1]{{#1}}
	
\fi

\newcommand{\comment}[1]{}

\newcommand{\rpm}{\raisebox{.2ex}{$\scriptstyle\pm$}}
\newcommand{\tabref}{Table~\ref}
\newcommand{\figref}{Fig.~\ref}
\newcommand{\figsref}{Figs.~\ref}
\newcommand{\secref}{Sec.~\ref}

\usepackage[breaklinks=true,bookmarks=false]{hyperref}

\iccvfinalcopy 

\def\iccvPaperID{226} 

\ificcvfinal\fi

\begin{document}

\title{Point-to-Point Video Generation}

\comment{
\author{Tsun-Hsuan Wang\footnotemark[1]\\
National Tsing Hua University\\
{\tt\small johnsonwang0810@gmail.com}
\and
Yen-Chi Cheng$^{*}$\\
National Tsing Hua University\\
{\tt\small charlescheng0117@gmail.com}
\and
Chieh Hubert Lin\\
National Tsing Hua University\\
{\tt\small hubert052702@gmail.com}
\and
Hwann-Tzong Chen\\
National Tsing Hua University\\
{\tt\small htchen@cs.nthu.edu.tw}
\and
Min Sun\\
National Tsing Hua University\\
{\tt\small sunmin@ee.nthu.edu.tw}
}
}

\author{%
Tsun-Hsuan Wang\footnotemark[1] , Yen-Chi Cheng$^{*}$, Chieh Hubert Lin, Hwann-Tzong Chen, Min Sun\\
National Tsing Hua University\\
{\tt\small \{johnsonwang0810, charlescheng0117, hubert052702\}@gmail.com}\\
{\tt\small htchen@cs.nthu.edu.tw}, {\tt\small sunmin@ee.nthu.edu.tw}
}

\twocolumn[{%
\maketitle
\renewcommand\twocolumn[1][]{#1}%
   \vspace{-5mm} 
    \centering
    \includegraphics[width=.94\linewidth]{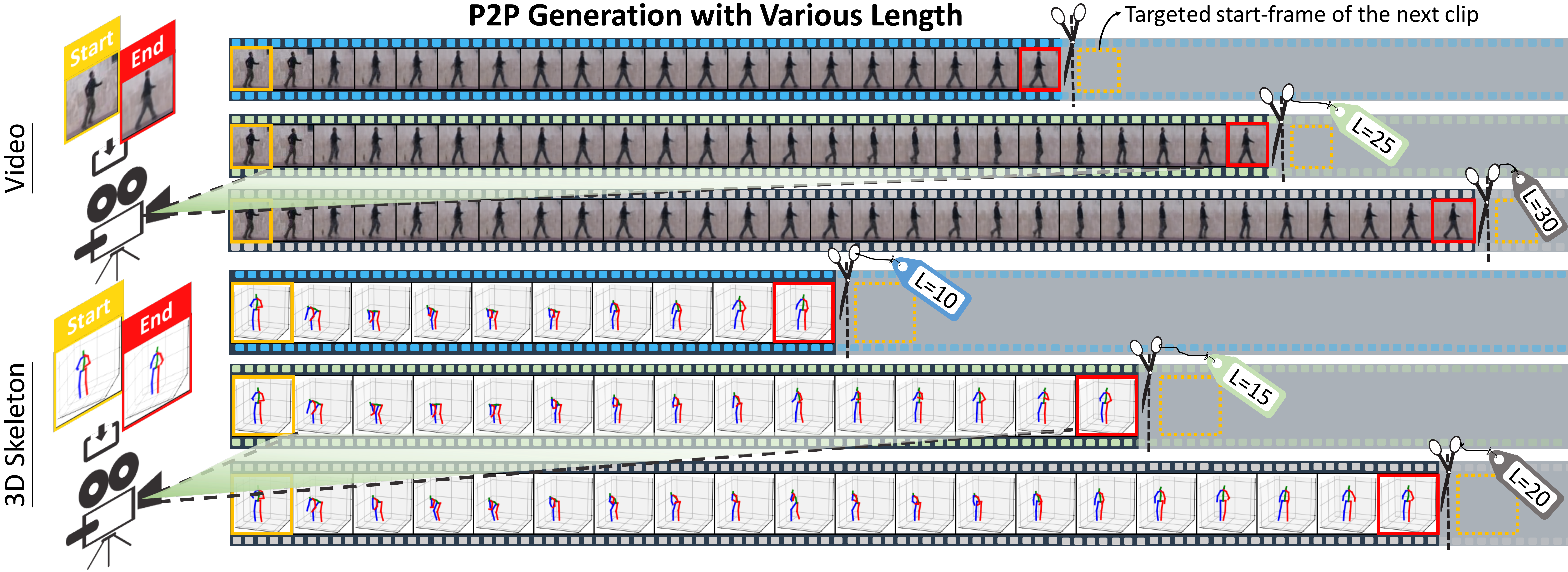}
    \vspace{-1.0em}
    \captionof{figure}{\textbf{Point-to-Point (P2P) Video Generation.} Given a pair of {(orange) start-} and {(red) end-}frames in the video and 3D skeleton domains, our method generates videos with smooth transitional frames of various lengths. The superb controllability of p2p generation naturally facilitates the modern video editing process.
    \label{teaser}\vspace{-0.5em}}
    \vspace{5mm} 
}]

\ificcvfinal\thispagestyle{empty}\fi

\begin{abstract} \vspace*{-1em}
    \footnotetext[1]{indicates equal contribution}
    While image synthesis achieves tremendous breakthroughs (\eg, generating realistic faces), video generation is less explored and harder to control, which limits its applications in the real world. For instance, video editing requires temporal coherence across multiple clips and thus poses both start and end constraints within a video sequence. We introduce \textbf{point-to-point video generation} that controls the generation process with two control points: the targeted start- and end-frames. The task is challenging since the model not only generates a smooth transition of frames, but also plans ahead to ensure that the generated end-frame conforms to the targeted end-frame for videos of various lengths. We propose to maximize the modified variational lower bound of conditional data likelihood under a skip-frame training strategy. Our model can generate end-frame-consistent sequences without loss of quality and diversity. We evaluate our method through extensive experiments on Stochastic Moving MNIST, Weizmann Action, Human3.6M, and BAIR Robot Pushing under a series of scenarios. The qualitative results showcase the effectiveness and merits of point-to-point generation. 
    \vspace*{-2.0em}
\end{abstract}

\section{Introduction}
\vspace{-0.4em}
The significant advancements in deep generative models bring impressive results in a wide range of domains such as image synthesis, text generation, and video prediction. Despite the huge success, unconstrained generation is still a few steps away from practical applications since it lacks intuitive and handy mechanisms to incorporate human manipulation into the generation process. In view of this incapability, conditional and controllable generative models have received an increasing amount of attention. Most existing work achieves controllability by conditioning the generation on the attribute, text, user inputs or scene graph \cite{johnson2018image,xu2018attngan,yan2016attribute2image,zhu2016generative}. However, regardless of the considerable progress in still image generation, controllable video generation is yet to be well explored.

Typically, humans create a video through breaking down the entire story into separate scenes, taking shots for each scene individually, and finally merging every piece of footage to form the final edit. This requires a smooth transition across not only frames but also different video clips, posing constraints on both start- and end-frames within a video sequence so as to align with the preceding and subsequent context. We introduce \textit{point-to-point video generation} (p2p generation) that controls the generation process with two control points---the targeted start- and end-frames. Enforcing consistency on the two control points allows us to regularize the context of the generated intermediate frames, and it also provides a straightforward strategy for merging multiple videos. Moreover, in comparison with standard video generation setting \cite{srivastava2015unsupervised}, which requires a consecutive sequence for initial frames, p2p generation only needs a pair of individual frames. Such a setting is more accessible in real-world scenarios, \eg, generating videos from images with similar content crawled on the Internet. Finally, p2p generation is preferable to attribute-based methods for more sophisticated video generation tasks that involve hard-to-describe attributes. Attribute-based methods heavily depend on the available attributes provided in the datasets, whereas p2p generation can avoid the burden of collecting and annotating meticulous attributes.

Point-to-point generation has two major challenges: {\em i}) The control point consistency (CPC) should be achieved without the sacrifice of generation quality and diversity. {\em ii}) The generation with various lengths should all satisfy the control point consistency. Following the recent progress in video generation and future frame prediction, we introduce a global descriptor, which carries information about the targeted end-frame, and a time counter, which provides temporal hints for dynamic length generation to form a conditional variational encoder (CVAE \cite{sohn2015learning}). In addition, to balance between generation quality, diversity, and CPC, we propose to maximize the modified variational lower bound of conditional data likelihood. Besides, we inject an alignment loss to ensure the latent space in the encoder and decoder aligns with each other. We further present the skip-frame training strategy to reinforce our model to be more time-counter-aware. Our model adjusts its generation procedure accordingly, and thus achieves better CPC. Extensive experiments are conducted on Stochastic Moving MNIST (or SM-MNIST) \cite{srivastava2015unsupervised,denton2018stochastic}, Weizmann Human Action \cite{ActionsAsSpaceTimeShapes_pami07}, and Human3.6M (3D skeleton data) \cite{h36m_pami} to evaluate the effectiveness of the proposed method. A series of qualitative results further highlight the merits of p2p generation and the capability of our model. \vspace{-0.6em}

\section{Related Work}
\vspace{-0.6em}

Our problem is most related to video generation~\cite{ranzato2014video,tulyakov2018mocogan,vondrick2016generating} and the controllability of video generation \cite{ hao2018controllable,he2018probabilistic,hu2018video,li2017video, marwah2017attentive,yamamoto2018conditional}. It also has a connection with video interpolation. We briefly review these topics in this section. 
\vspace{-1.2em}

\paragraph{Video Generation.} 
Many approaches use GANs \cite{aigner2018futuregan,tulyakov2018mocogan,vondrick2016generating} or adversarial loss during training for generating videos \cite{aigner2018futuregan,liang2017dual,lotter2016unsupervised,mathieu2016deep,saito2017temporal,vondrick2016generating,vondrick2017generating}. Vondrick~\etal \cite{vondrick2016generating} use a generator with two pathways to predict the foreground and background, and a discriminator to distinguish a video as real or fake. On the other hand, it can be tackled by learning how to transform observed frames to synthesize the future frames \cite{finn2016unsupervised,jia2016dynamic,liu2017video,vondrick2017generating,xue2016visual}.
Furthermore, strategies based on decomposing a video into a static part that can be shared along (\ie content) and the varying part (\ie motion) are also proposed to describe the video dynamics  \cite{denton2017unsupervised,hsieh2018learning,tulyakov2018mocogan,villegas2017decomposing,wiles2018self}. Denton~\etal \cite{denton2017unsupervised} encode motion and content into different subspaces and use an adversarial loss on the motion encoder to achieve disentanglement.



Several methods rely on VAE \cite{kingma2013auto} to capture the uncertain nature in videos \cite{babaeizadeh2018stochastic,denton2018stochastic, fragkiadaki2017motion,he2018probabilistic,lee2018stochastic,liang2017dual,walker2017pose,yan2018mt}. Babaeizadeh~\etal \cite{babaeizadeh2018stochastic} extend \cite{finn2016unsupervised} with variational inference framework such that their model can predict multiple frames of plausible futures on real-world data. Jayaraman~\etal \cite{jayaraman19time} predicts the most certain frame first and breaks down the original problem such that the predictor can complete the semantic sub-goals coherently.

While the methods mentioned above achieve good results on video prediction, the generation process is often uncontrollable and hence leads to unconstrained outputs. In order to preserve the ability of generating diversified outputs while achieving control point consistency, we manage to build upon VAE for point-to-point video generation.
\vspace{-1.2em}

\paragraph{Video Interpolation (VI).}
\label{sec:vi_paragraph}
\charles{The problem setting of p2p generation has connection with VI~\cite{liu2017voxelflow,Niklaus_CVPR_2017,Niklaus_ICCV_2017,meyer2018phasenet,jiang2018super}, but with essential difference. VI aims to increase the frame-rate of a video. Thus both the number of inserted frames and the time interval of interpolation are assumed to be small, whereas p2p generation involves a much longer-term synthesis of in-between frames, posing a different challenge. Besides, VI methods typically are deterministic (i.e., producing only one interpolated result). Instead, our work is closer to video generation where the synthesized frames are required to be both \textit{temporally coherent} and \textit{diverse} in context. Finally, automatic looping (i.e., generating a looping video given the same start and end-frame) can be accomplished by p2p generation but not by VI (see \secref{ssec:vi_comparison} for detailed analysis).}
\vspace{-1.2em}

\paragraph{Controllability on Video Generation.} 
Several methods are proposed to guide the video generation process. Hu~\etal \cite{hu2018video} use an image and a motion stroke to synthesize the video. Hao~\etal \cite{hao2018controllable} conditions on the start frame and a trajectory provided by user to steer the motion and appearance for the next frames. He~\etal \cite{he2018probabilistic} proposes an attribute-based approach for transient control by exploiting the attributes (\eg, identity, action) in the dataset. Text or language features can also be used as the instruction for controls \cite{li2017video,marwah2017attentive,yamamoto2018conditional}. Although the existing methods all provide freedom for controlling the generation, they come with some limitations. Conditioning on language would suffer from its ambiguous nature, which does not allow precise control \cite{marwah2017attentive}. Attribute control depends on the data labels and is not be available in unsupervised setting. User provided input is intuitive but requires annotations during training. Instead, our method {\em i}) only conditions on the target frame which can be acquired without any cost, {\em ii}) can incorporate with detailed description of the control points (\eg, the precise look and action of a person, or joints of a skeleton) to provide exact control, {\em iii}) can be trained in a fully unsupervised fashion. The advantage over previous methods in having the controllability of start- and target-frames motivates our point-to-point generation.
\vspace{-0.6em} 

\section{Methodology}
\vspace{-0.6em} 
\begin{figure*}[t!]
    \centering
    \includegraphics[width=\linewidth]{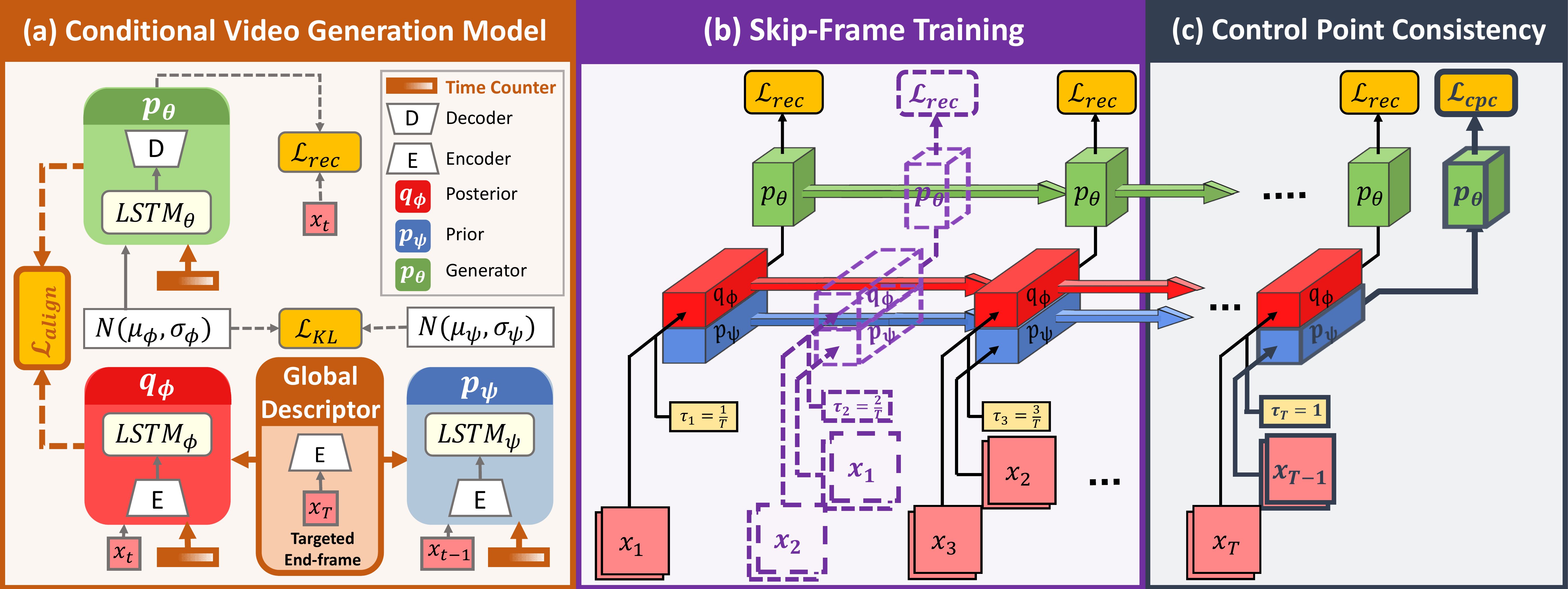} \vspace{-1.5em}
    \caption{\textbf{An overview of the novel components of p2p generation.} (a) Our model is a VAE consisting of posterior $q_\phi$, prior $p_\psi$, and generator $p_\theta$ (all with an LSTM for temporal coherency). We use KL-divergence to encourage $p_\psi$ to be similar to $q_\phi$. To control the generation, we encode the targeted end-frame $x_T$ into a global descriptor. Both $q_\phi$ and $p_\psi$ are computed by considering not only the input frame ($x_t$ or $x_{t-1}$), but also the ``global descriptor" and ``time counter". We further use the ``alignment loss'' to align the encoder and decoder latent space to reinforce the control point consistency. (b) Our skip-frame training has a probability to skip the input frame in each timestamp where the input will be ignored completely and the hidden state will not be propagated at all \textbf{(see the dashed line)}. (c) The control point consistency is achieved by posing CPC loss on $p_\psi$ without harming the reconstruction objective of $q_\phi$ \textbf{(highlighted in bold)}.}
    \label{fig:overview}
    \vspace{-1.5em}
\end{figure*}

Given a pair of control points (the targeted start- and end-frames $\{x_1, x_T\}$) and the generation length $T$, we aim to generate a sequence $\hat{x}_{1:T}$ with the specified length such that their start- and end-frames $\{\hat{x}_1, \hat{x}_T\}$ are consistent with the control points. To maintain quality and diversity in p2p generation, we present a conditional video generation model (\secref{ssec:global_and_counter}) that maximizes the modified variational lower bound (\secref{ssec:cpc_on_prior}). To further improve CPC under various lengths, we propose a novel skip-frame training strategy (\secref{ssec:skip_frame}) and a latent alignment loss (\secref{ssec:final_obj}).
\vspace{-0.4em}

\subsection{VAE and Video Generation}
\label{ssec:base_model}
\vspace{-0.5em}
Variational Autoencoder (VAE) leverages a simple prior $p_\theta(\mathbf{z})$ (\eg, Gaussian) and a complex likelihood $p_\theta(\mathbf{x}|\mathbf{z})$ (\eg, a neural network) on latent variable $\mathbf{z}$ to maximize the data likelihood $p_\theta(\mathbf{x})$, where $\mathbf{x}=[x_1, x_2,\dots, x_T]$. A variational neural network $q_\phi(\mathbf{z}|\mathbf{x})$ is introduced to approximate the intractable latent posterior $p_\theta(\mathbf{z}|\mathbf{x})$, allowing joint optimization over $\theta$ and $\phi$,
\vspace{-0.6em}
\begin{equation}
	\begin{split}
	\log p_\theta(\mathbf{x}) &= \log \int_{\mathbf{z}}p_\theta(\mathbf{x}|\mathbf{z})p(\mathbf{z}) \, d\mathbf{z}\\
	& \hspace{-2em} \geq \mathbb{E}_{q_\phi(\mathbf{z}|\mathbf{x})}\log p_\theta(\mathbf{x}|\mathbf{z}) - D_\mathrm{KL}(q_\phi(\mathbf{z}|\mathbf{x})||p(\mathbf{z})) \,.
    \end{split}
    \label{eq:vae}
    \vspace{-0.6em}
\end{equation}
\noindent The intuition behind the inequality is to reconstruct data $\mathbf{x}$ with latent variable $\mathbf{z}$ sampled from the posterior $q_\phi(\mathbf{z}|\mathbf{x})$, simultaneously minimizing the KL-divergence between the prior $p(\mathbf{z})$ and posterior $q_\phi(\mathbf{z}|\mathbf{x})$.

Video generation commonly adopts VAE framework accompanied by a recurrent model (\eg, LSTM), where the VAE handles generation process and the recurrent model captures the dynamic dependencies in sequential generation. However, in VAE, the simple choice for prior $p(\mathbf{z})$ such as a fixed Gaussian $\mathcal{N}(\mathbf{0},\mathbf{I})$ is confined to drawing samples randomly at each timestep regardless of temporal dependencies across frames. Accordingly, existing works resort to parameterizing the prior with a learnable function $p_\psi(z_t|x_{1:t-1})$ conditioned on previous frames $x_{1:t-1}$. The variational lower bound throughout the entire sequence is \vspace{-0.5em}
\begin{equation}
	\begin{split}
	\mathcal{L}_{\theta,\phi,\psi}(x_{1:T}) & =  \sum_{t=1}^{T}\big[ \, \mathbb{E}_{q_{\phi}(z_{1:t}|x_{1:t})}\log p_\theta(x_t|x_{1:t-1},z_{1:t}) \\
	 - & D_\mathrm{KL}(q_{\phi}(z_{1:t}|x_{1:t})||p_\psi(z_t|x_{1:t-1})) \, \big] \, . \hspace{1.5em}
	\end{split}
    \label{eq:svg}
\end{equation}
%
In comparison with a standard VAE, the former term describes the reconstruction sampled from the posterior $q_\phi(z_{1:t}|x_{1:t})$ conditioned on data up to the \textit{current frame}. The latter term ensures that the prior $p_\psi(z_t|x_{1:t-1})$ conditioned on data up to the \textit{previous frame} does not deviate from the posterior. Meanwhile, it also serves as a regularization on the learning of posterior. In this work, we inherit and modify the network architecture of \cite{denton2018stochastic} and adapt $\mathcal{L}_{\theta,\phi,\psi}(x_{1:T})$ for p2p generation.
\vspace{-0.3em}

\subsection{Global Descriptor and Time Counter}
\label{ssec:global_and_counter}
\vspace{-0.5em}
For a deep network to achieve p2p generation under various lengths, {\em i}) the model should be aware of the information of control points and {\em ii}) the model should be able to perceive time lapse and generate the targeted end-frame at the designated timestep. While the targeted start-frame is already fed as an initial frame, we adopt a straightforward strategy to incorporate the control points into the model at every timestep by feeding features encoded from the targeted end-frame $h_T$ to our model. Besides, to enforce our model to be aware of when to generate the targeted end-frame given the generation length $T$, we introduce a time counter $\tau_t\in [0,1]$, where $\tau_t=0.0$ indicates the beginning of the sequence and $\tau_t=1.0$ indicates reaching the targeted end-frame. As shown in \figref{fig:overview}(a), $q_\phi$ and $p_\psi$ are modeled by a shared-weight encoder and two different LSTMs, and $p_\theta$ is modeled by the third LSTM along with a decoder to map latent vectors to image space. The inference process during training at timestep $t$ is shown as
\vspace{-0.3em}
\begin{equation}
    \begin{split}
        h_T &= Enc(x_T),~~~~\tau_t=t/T, \\
        \mu^{t}_\phi, \sigma^{t}_\phi &= \mathrm{LSTM}_\phi(h_t, h_T, \tau_t), \; h_t=Enc(x_t), \\
        z_t &\sim \mathcal{N}(\mu^{t}_\phi, \sigma^{t}_\phi), \\
        g_t &= \mathrm{LSTM}_\theta(h_{t-1}, z_t, \tau_t), \; h_{t-1}=Enc(x_{t-1}), \\
        \hat{x}_t &= Dec(g_t) \,.
    \end{split}
    \label{eq:infer_training}
    \vspace{-0.7em}
\end{equation}
\noindent During test time, as we have no access to current $x_t$, the latent variable $z_t$ is sampled from the prior distribution $p_\psi$,
\vspace{-0.5em}
\begin{equation}
    \begin{split}
        \mu^{t}_\psi, \sigma^{t}_\psi &= \mathrm{LSTM}_\psi(h_{t-1}, h_T, \tau_t), \\
        z_t &\sim \mathcal{N}(\mu^{t}_\psi, \sigma^{t}_\psi) \,. 
    \end{split}
    \label{eq:infer_testing}
    \vspace{-1.0em}
\end{equation}
\noindent Recall that the KL divergence in \eqref{eq:svg} enforces the alignment between $q_\phi$ and $p_\psi$, allowing $p_\psi$ to serve as a proxy of $q_\phi$ at test time. Besides, by introducing the global descriptor $h_T$ and time counter $\tau_t$, \eqref{eq:svg} is extended to a variational lower bound of conditional data likelihood $\mathcal{L}_{\theta,\phi,\psi}(x_{1:T}|\mathbf{c})$, where $\mathbf{c}$ is the conditioning on the targeted end-frame and time counter. In addition, we further propose a latent space alignment loss within $h_t$ and $g_t$ to mitigate the mismatch between encoding and decoding process, as shown in \eqref{eq:final_obj}.

\subsection{Control Point Consistency on Prior}
\label{ssec:cpc_on_prior}
\vspace{-0.4em}
Although introducing the time counter and the global descriptor of control points provides the model with capability of achieving CPC, we are not able to further reinforce the generated end-frame to conform to the targeted end-frame. While the conditioning happens to be a part of the reconstruction objective, naively increasing the weight $\alpha_\mathrm{cpc}$ at timestep $T$ in the reconstruction term of \eqref{eq:svg}, \ie, $\sum_{t=1}^{T-1}\mathbb{E}_{q_\phi}\log p_\theta(x_t)+\alpha_\mathrm{cpc}\mathbb{E}_{q_\phi}\log p_\theta(x_T)$, results in unstable training behavior and degradation of generation quality and diversity. To tackle this problem, we propose to single out the CPC from the reconstruction loss on the posterior and pose it on the prior. The modified lower bound of conditional data likelihood with a learnable prior $p_\psi$ is
\vspace{-0.6em}
\begin{equation}
	\begin{split}
	\mathcal{L}_{\theta,\phi,\psi}^\mathrm{p2p} & (x_{1:T}|\mathbf{c})  = \\  \sum_{t=1}^{T} & \big[\mathbb{E}_{q_{\phi}(z_{1:t}|x_{1:t},\mathbf{c})}\log p_{\theta}(x_t|x_{1:t-1},z_{1:t},\mathbf{c}) \\
	 - &~ D_\mathrm{KL}(q_{\phi}(z_{1:t}|x_{1:t},\mathbf{c})||p_\psi(z_t|x_{1:t-1},\mathbf{c})) \big] \\
	 + &~ \mathbb{E}_{p_{\psi}(z_T|x_{1:T-1},\mathbf{c})}\log p_{\theta}(x_T|x_{1:T-1},z_{1:T},\mathbf{c}) \,. 
    \end{split}
    \label{eq:p2p_vae}
    \vspace{-0.6em}
\end{equation}
\noindent While the first two terms are the same as the bound of conditional VAE (CVAE), the third term of the above formulation benefits a more flexible tuning on the behavior of the additionally-introduced condition without degrading the maximum likelihood estimate in the first term. 

\subsection{Skip-Frame Training}
\label{ssec:skip_frame}
\vspace{-0.6em}
A well-functioning p2p generation model should be aware of the time counter in order to achieve CPC under various lengths. However, most video datasets have a fixed frame rate. As a result, the model may exploit the fixed frequency across frames and ignore the time counter. We introduce skip-frame training to further enhance the model to be more aware of the time counter. Basically, we randomly drop frames while computing the reconstruction loss and KL divergence (the first two terms in \eqref{eq:p2p_vae}). The LSTMs are hence forced to take time counter into consideration so as to handle the random skipping in the recurrence. Such adaption in the maximum likelihood estimate of posterior $q_\phi$ further incorporates CPC into the learning of posterior.

\subsection{Final Objective}
\label{ssec:final_obj}
\vspace{-0.6em}
To summarize, our final objective that maximizes the modified variational lower bound of conditional data likelihood under a skip-frame training strategy is
\vspace{-0.2em}
\begin{equation}
    \small
	\begin{split}
	& \hspace{-0.8em} \mathcal{L}_{\theta,\phi,\psi}^\mathrm{full}(x_{1:T}|\mathbf{c}) = \\
	& \hspace{1em} \sum_{t=1}^{T}M_t\Big[\mathbb{E}_{q_{\phi}(z_{1:t}|x_{1:t},\mathbf{c})}\log p_\theta(x_t|x_{1:t-1},z_{1:t},\mathbf{c})  \\
	& \hspace{4em}-\beta D_\mathrm{KL}(q_{\phi}(z_{1:t}|x_{1:t},\mathbf{c})||p_\psi(z_t|x_{1:t-1},\mathbf{c})) \\
	& \hspace{4em}-\alpha_\mathrm{align} ||h_t-g_t||_2\Big] \\
	& \hspace{1em} +\alpha_\mathrm{cpc} \mathbb{E}_{p_{\psi}(z_T|x_{1:t-1},\mathbf{c})}\log p_\theta(x_T|x_{1:T-1},z_{1:T},\mathbf{c}), \\
    \end{split}
    \label{eq:final_obj}
\end{equation}
\noindent where $ M_t \sim \mathrm{Bernoulli}(1-p_\mathrm{skip})$, $M_T=1$. $\beta$, $\alpha_\mathrm{cpc}$, and $\alpha_\mathrm{align}$ are hyperparameters to balance between KL term, CPC, and latent space alignment. The constant $p_\mathrm{skip} \in [0,1]$ determines the rate of skip-frame training. 

\section{Experiment}
\vspace{-0.15em} 

\begin{table*}[t]
\centering
\small
\setlength\tabcolsep{4.5pt}
\begin{tabular}{l | c c c c | c c c c}
\hline
\multirow{2}{*}{Method} & \multicolumn{4}{c|}{SSIM (\rpm~indicates $95\%$ confidence interval)}                                                                                 & \multicolumn{4}{c}{PSNR (\rpm~indicates $95\%$ confidence interval)} \\ 
                  & S-Best $\shortuparrow$                   & S-Div (1E-3) $\shortuparrow$              & S-CPC $\shortuparrow$                     & R-Best $\shortuparrow$                    & S-Best $\shortuparrow$                     & S-Div $\shortuparrow$                     & S-CPC $\shortuparrow$                      & R-Best $\shortuparrow$ \\ \hline \hline \\ [-2.3ex]
SVG \cite{denton2018stochastic} & $0.780^{\text{\rpm}0.006}$ & $2.349^{\text{\rpm}0.076}$ & $0.621^{\text{\rpm}0.004}$ & $0.850^{\text{\rpm}0.005}$ & $15.774^{\text{\rpm}0.161}$ & $0.816^{\text{\rpm}0.019}$ & $12.105^{\text{\rpm}0.047}$ & $18.001^{\text{\rpm}0.201}$ \\
+ C     & $0.768^{\text{\rpm}0.002}$ & $2.482^{\text{\rpm}0.048}$ & $0.729^{\text{\rpm}0.003}$ & $0.840^{\text{\rpm}0.004}$ & $15.373^{\text{\rpm}0.049}$ & $0.914^{\text{\rpm}0.014}$ & $14.024^{\text{\rpm}0.054}$ & $17.751^{\text{\rpm}0.094}$ \\
+ C + A & $0.755^{\text{\rpm}0.003}$ & $2.377^{\text{\rpm}0.085}$ & $0.735^{\text{\rpm}0.005}$ & $0.816^{\text{\rpm}0.005}$ & $15.117^{\text{\rpm}0.103}$ & $0.804^{\text{\rpm}0.014}$ & $14.141^{\text{\rpm}0.069}$ & $16.884^{\text{\rpm}0.147}$ \\
Ours             & $0.755^{\text{\rpm}0.004}$ & $2.525^{\text{\rpm}0.052}$ & $\mathbf{0.769}^{\text{\rpm}0.005}$ & $0.832^{\text{\rpm}0.005}$ & $15.265^{\text{\rpm}0.079}$ & $0.815^{\text{\rpm}0.009}$ & $\mathbf{15.185}^{\text{\rpm}0.096}$ & $17.581^{\text{\rpm}0.172}$ \\
\hline
\end{tabular} \vspace{-0.5em}
\caption{Evaluation on SM-MNIST (\textit{+C}: CPC loss on $p_\psi$ only. \textit{+C+A}: CPC loss and Alignment loss. \textit{Ours}: Our full model).}
\label{tab:quantitative_mnist}
\vspace{-1.0em} 
\end{table*}

\begin{table*}[t]
\centering
\small
\setlength\tabcolsep{4.5pt}
\begin{tabular}{l | c c c c | c c c c}
\hline
\multirow{2}{*}{Method} & \multicolumn{4}{c|}{SSIM (\rpm~indicates $95\%$ confidence interval)}                                                                                  & \multicolumn{4}{c}{PSNR (\rpm~indicates $95\%$ confidence interval)}                                                                       \\ 
                 & S-Best $\shortuparrow$                    & S-Div (1E-3) $\shortuparrow$              & S-CPC $\shortuparrow$                      & R-Best $\shortuparrow$                     & S-Best $\shortuparrow$                     & S-Div $\shortuparrow$                     & S-CPC $\shortuparrow$                      & R-Best $\shortuparrow$ \\ \hline \hline \\ [-2.3ex]
SVG \cite{denton2018stochastic} & $0.819^{\text{\rpm}0.008}$ & $1.992^{\text{\rpm}0.351}$ & $0.734^{\text{\rpm}0.008}$ & $0.819^{\text{\rpm}0.009}$ & $25.234^{\text{\rpm}0.355}$ & $1.904^{\text{\rpm}0.357}$ & $22.236^{\text{\rpm}0.242}$ & $25.039^{\text{\rpm}0.400}$ \\
+ C     & $0.814^{\text{\rpm}0.005}$ & $2.574^{\text{\rpm}0.402}$ & $0.730^{\text{\rpm}0.004}$ & $0.808^{\text{\rpm}0.006}$ & $24.898^{\text{\rpm}0.110}$ & $2.186^{\text{\rpm}0.346}$ & $22.028^{\text{\rpm}0.084}$ & $24.624^{\text{\rpm}0.211}$ \\
+ C + A & $0.823^{\text{\rpm}0.005}$ & $1.225^{\text{\rpm}0.178}$ & $0.767^{\text{\rpm}0.009}$ & $0.822^{\text{\rpm}0.005}$ & $25.092^{\text{\rpm}0.186}$ & $1.266^{\text{\rpm}0.170}$ & $22.855^{\text{\rpm}0.197}$ & $24.848^{\text{\rpm}0.145}$ \\
Ours             & $0.824^{\text{\rpm}0.004}$ & $1.106^{\text{\rpm}0.078}$ & $\mathbf{0.783}^{\text{\rpm}0.003}$ & $0.842^{\text{\rpm}0.006}$ & $24.993^{\text{\rpm}0.103}$ & $1.039^{\text{\rpm}0.057}$ & $\mathbf{23.334}^{\text{\rpm}0.105}$ & $25.660^{\text{\rpm}0.154}$ \\
\hline
\end{tabular} \vspace{-0.5em}
\caption{Evaluation on Weizmann (\textit{+C}: CPC loss on $p_\psi$ only. \textit{+C+A}: CPC loss and Alignment loss. \textit{Ours}: Our full model).}
\label{tab:quantitative_weizmann}
\vspace{-1.0em}
\end{table*}

\begin{table}[t]
\centering
\small
\begin{tabular}{l | c c c c}
\hline
Method          & S-Best $\shortdownarrow$           & S-Div $\shortuparrow$                   & S-CPC $\shortdownarrow$         & R-Best $\shortdownarrow$ \\ \hline \hline \\ [-2.3ex]
SVG \cite{denton2018stochastic} & $6.49^{\text{\rpm}0.31}$ & $0.68^{\text{\rpm}0.05}$ & $10.83^{\text{\rpm}0.90}$ & $5.75^{\text{\rpm}0.17}$ \\
+ C             & $8.25^{\text{\rpm}0.65}$ & $0.64^{\text{\rpm}0.06}$ & $12.08^{\text{\rpm}0.65}$ & $8.97^{\text{\rpm}0.53}$ \\
+ C + A  & $4.96^{\text{\rpm}0.18}$ & $0.80^{\text{\rpm}0.03}$ & $6.66^{\text{\rpm}0.82}$  & $4.74^{\text{\rpm}0.17}$ \\
Ours         & $4.46^{\text{\rpm}0.35}$ & $0.88^{\text{\rpm}0.06}$ & $\mathbf{0.72}^{\text{\rpm}0.06}$ & $1.23^{\text{\rpm}0.04}$ \\
\hline
\end{tabular}
\vspace{-0.8em}
\caption{Evaluation on Human3.6M (with MSE).}
\label{tab:quantitative_h36m}
\vspace{-1em}
\end{table}

\begin{table}[t]
\footnotesize
\setlength\tabcolsep{4.5pt}
\begin{tabular}{l | c c c c}
\hline
Method & S-Best $\shortuparrow$      & S-Div (1E-3) $\shortuparrow$ & S-CPC $\shortuparrow$      & R-Best $\shortuparrow$ \\ \hline \hline \\ [-2.3ex]
SVG~\cite{denton2018stochastic} & $0.845^{\text{\rpm}.006}$  & $0.716^{\text{\rpm}.166}$   & $0.775^{\text{\rpm}.008}$ & $0.926^{\text{\rpm}.003}$ \\
SV2P~\cite{babaeizadeh2018stochastic} & $0.841^{\text{\rpm}.010}$ & $0.186^{\text{\rpm}.021}$ & $0.770^{\text{\rpm}.009}$ & $0.847^{\text{\rpm}.004}$ \\
Ours & $0.847^{\text{\rpm}.004}$ & $0.664^{\text{\rpm}.049}$ & $\mathbf{0.824}^{\text{\rpm}.015}$ & $0.907^{\text{\rpm}.006}$ \\
\hline
\end{tabular}
\vspace{-0.8em}
\caption{\charles{Evaluation on BAIR Robot Pushing (with SSIM).}}
\label{tab:quantitative_bair}
\vspace{-1em} 
\end{table}

\begin{figure}[t!]
    \centering
    \includegraphics[width=\linewidth]{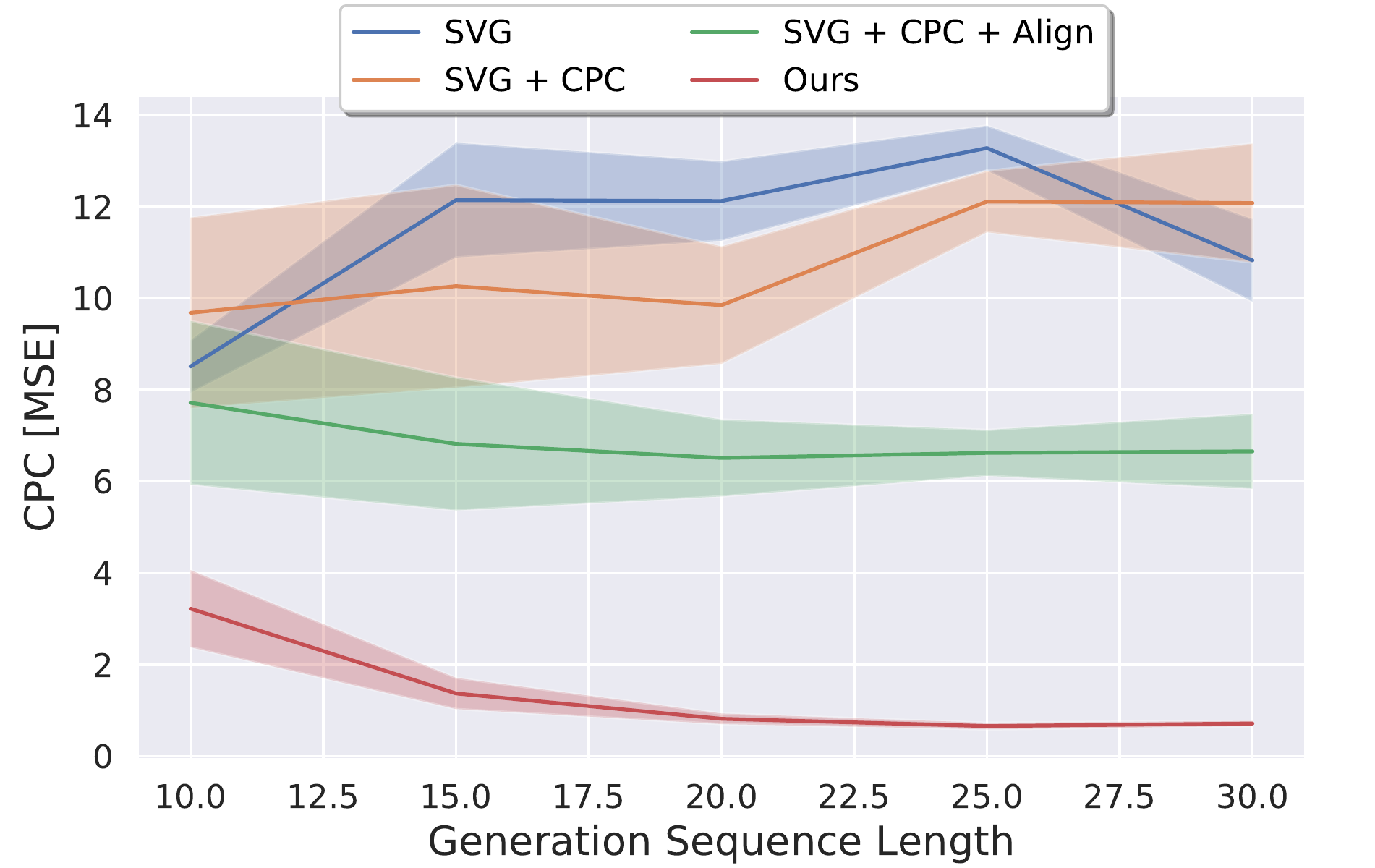} 
    \vspace{-1.5em}
    \caption{Control Point Consistency (CPC) with various generation lengths shows that our final model (in red) is more stable and can steadily approach the targeted end-frame \charles{(Figures are all best viewed in colors)}.} 
    \label{fig:dynlen} 
    \vspace{-1.0em}
\end{figure}

\begin{figure}[t]
    \centering
    \includegraphics[width=\linewidth]{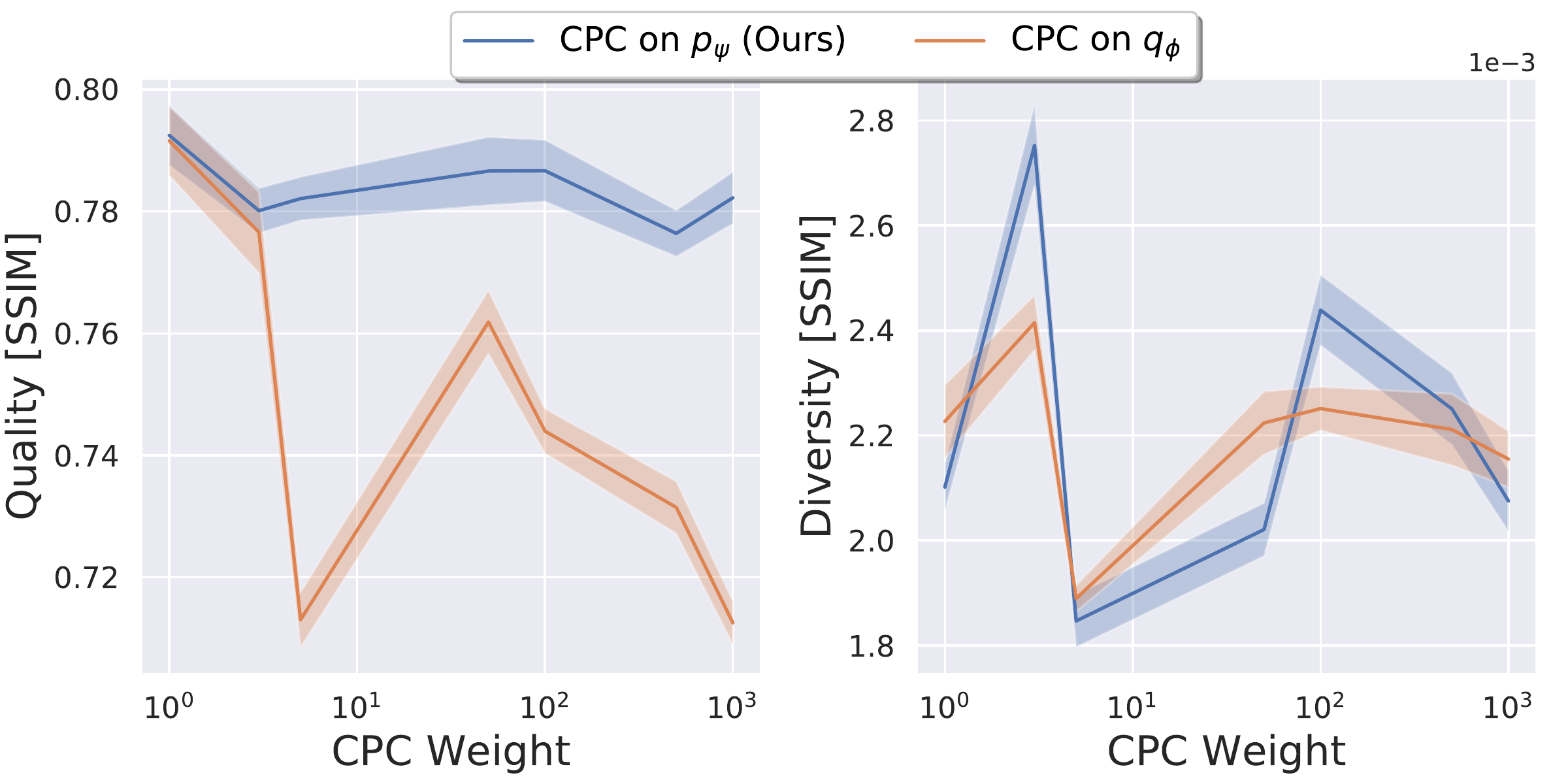} \vspace{-1.5em}
    \caption{We show the generation quality and diversity with different CPC weights. The results show that posing CPC on prior is more stable than on posterior; the latter is sensitive to large CPC weights and tends to harm the quality.}
    \label{fig:cpc_prior_good}
    \vspace{-1.0em}
\end{figure}

\begin{figure}[t!]
    \centering
    \includegraphics[width=\linewidth]{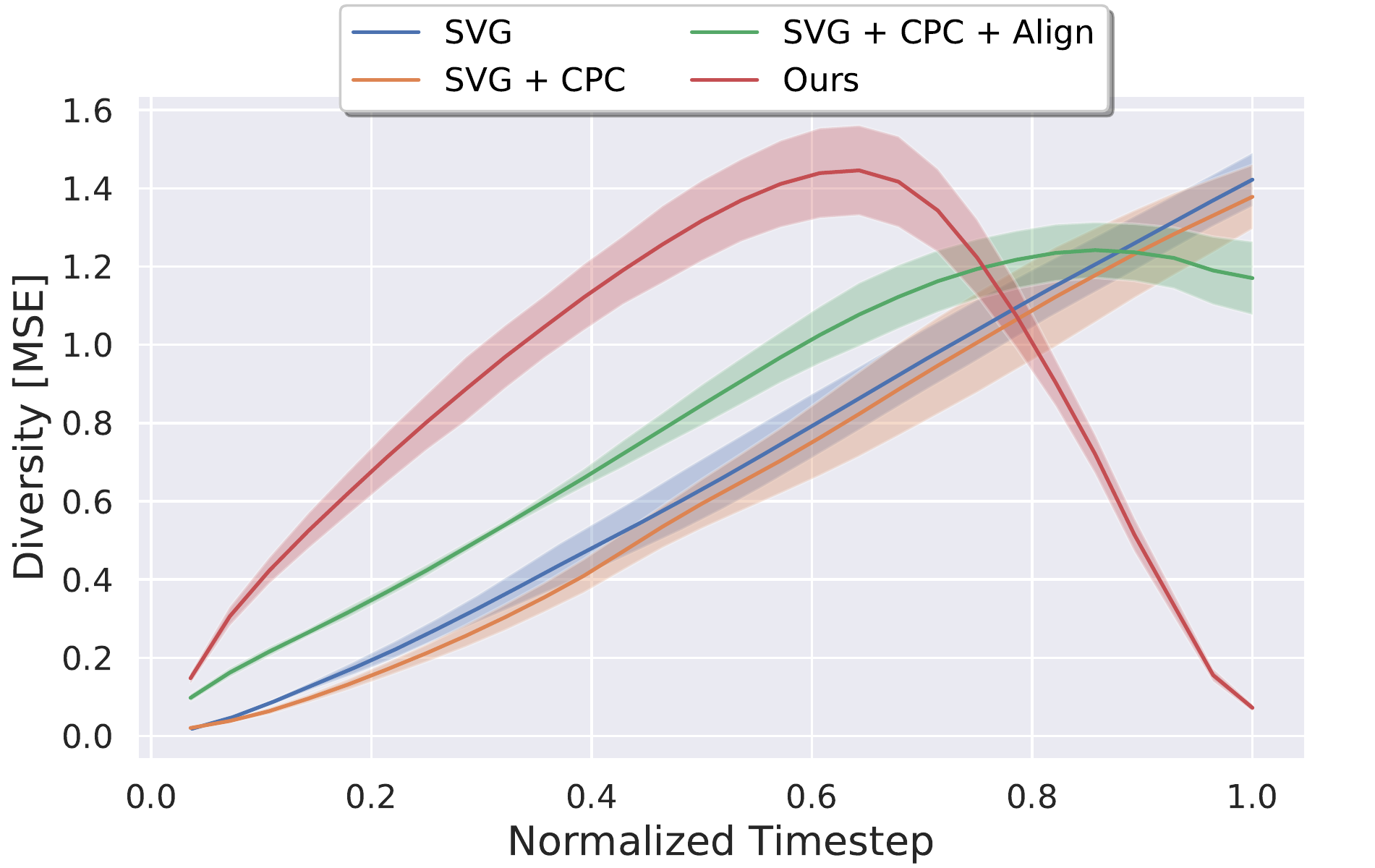} \vspace{-1.5em}
    \caption{The generation diversity through normalized time-steps shows that \textit{Ours} (in red) presents a desired behavior---diversity increases until the middle of generation, then converges (decreases) at targeted end-frames.}
    \label{fig:diversity}
    \vspace{-1.0em}
\end{figure}

\begin{figure}[h]
    \centering
    \captionsetup{font=small}
    \includegraphics[width=\linewidth]{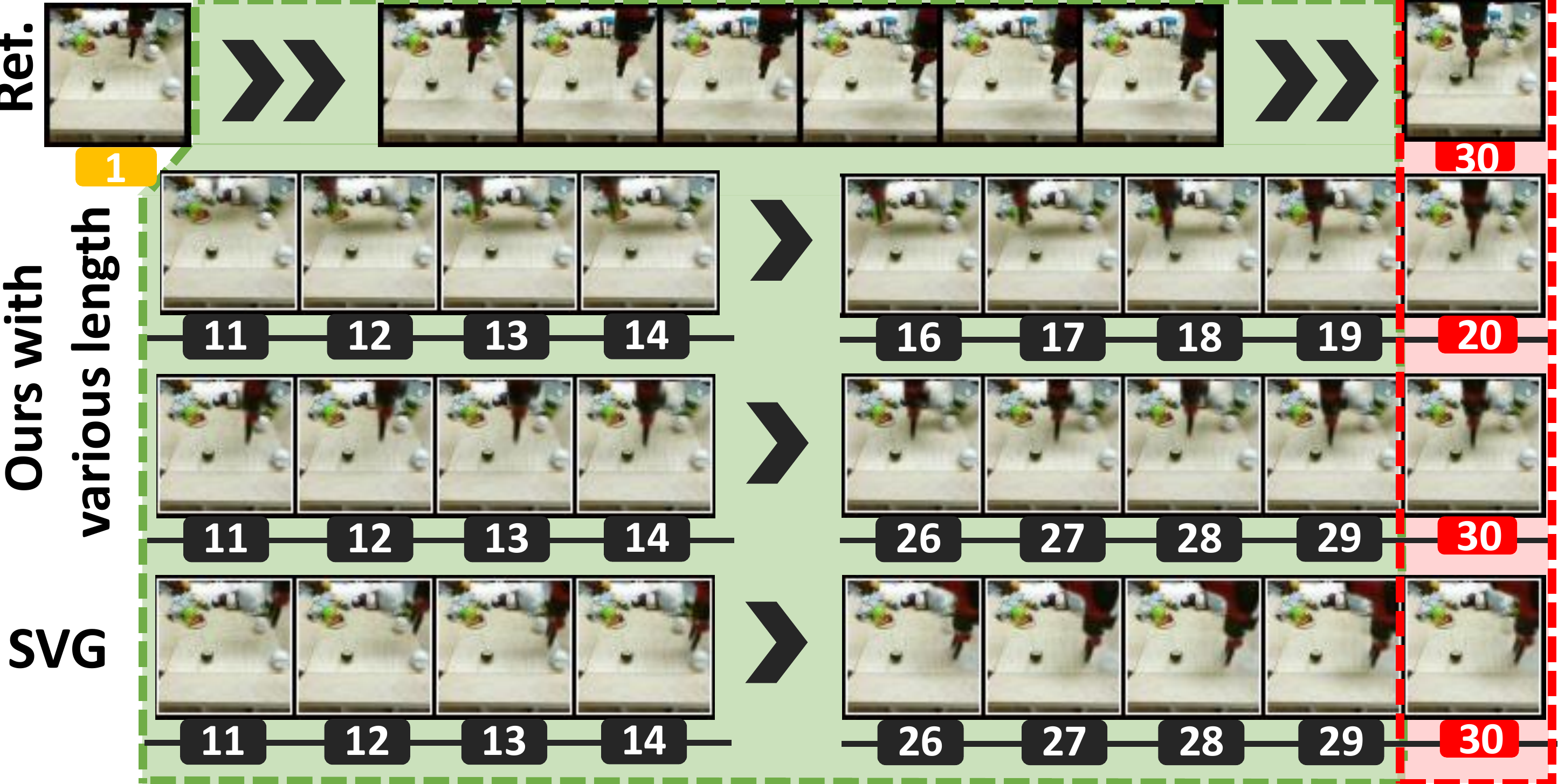}
    \vspace{-1.5em}
    \caption{Generation with various lengths on BAIR Pushing.}
    \label{fig:dynlen_bair}
    \vspace{-1.2em}
\end{figure}

\begin{figure*}[t!]
    \centering
    \includegraphics[width=\linewidth]{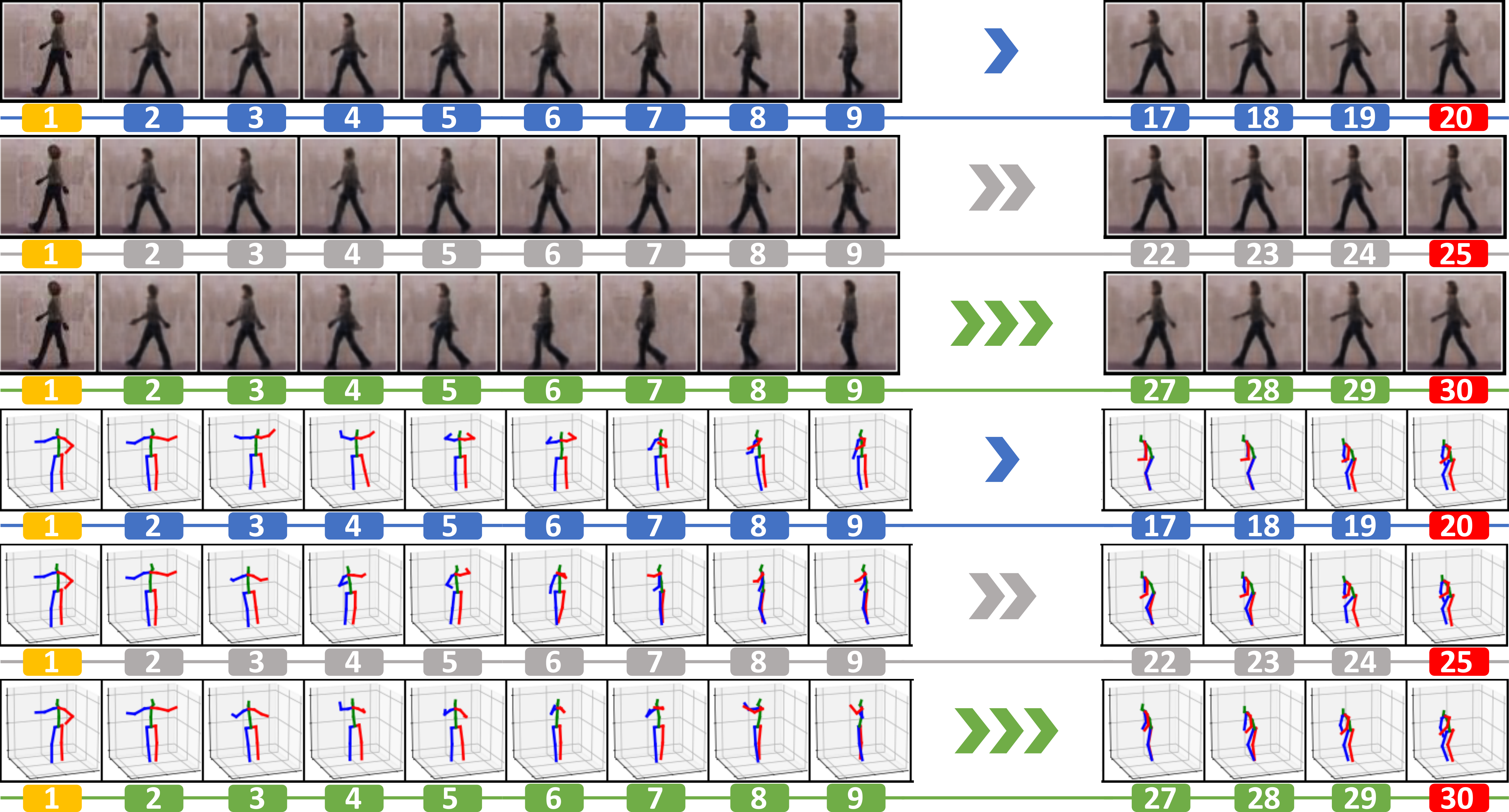}
    \vspace{-1.5em}
    \caption{Given a pair of {(orange) start-} and {(red) end-}frames, we show various lengths generation on Weizmann and Human3.6M (Number beneath each frame indicates the timestamp). Our model can achieve high-intermediate-diversity and targeted end-frame consistency while aware of various-length generation at the same time.}
    \label{fig:vary_len}
    \vspace{-0.8em}
\end{figure*}

\begin{figure*}[t!]
    \centering
    \includegraphics[width=\linewidth]{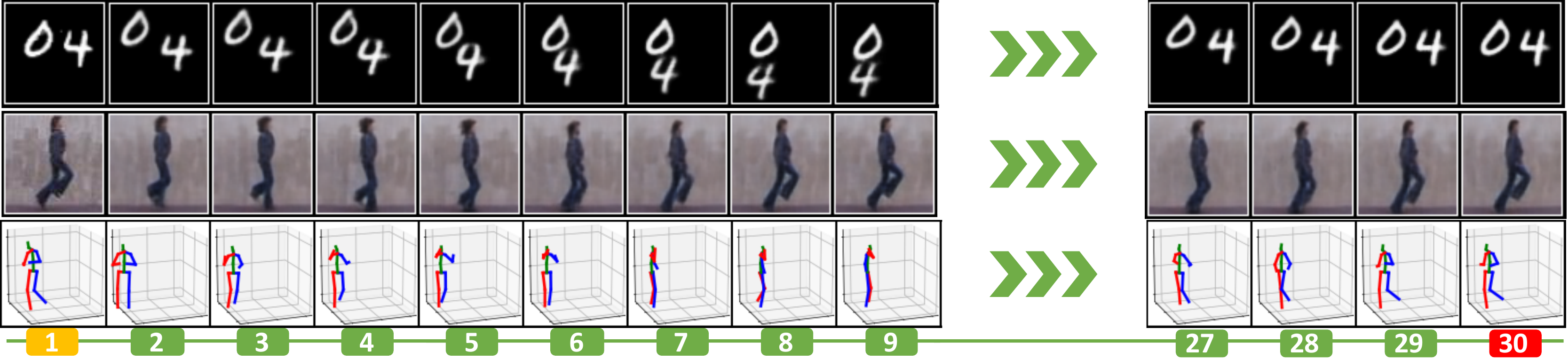}
    \vspace{-1.5em}
    \caption{We set the {(orange) start-} and {(red) end-}frame with the same frame to achieve loop generation. Our model can generate videos that form infinite loops while preserving diversity. See more results in supplementary materials.}
    \label{fig:loop_gen}
    \vspace{-1.0em}
\end{figure*}

\begin{figure*}[h]
    \centering
    \includegraphics[width=\linewidth]{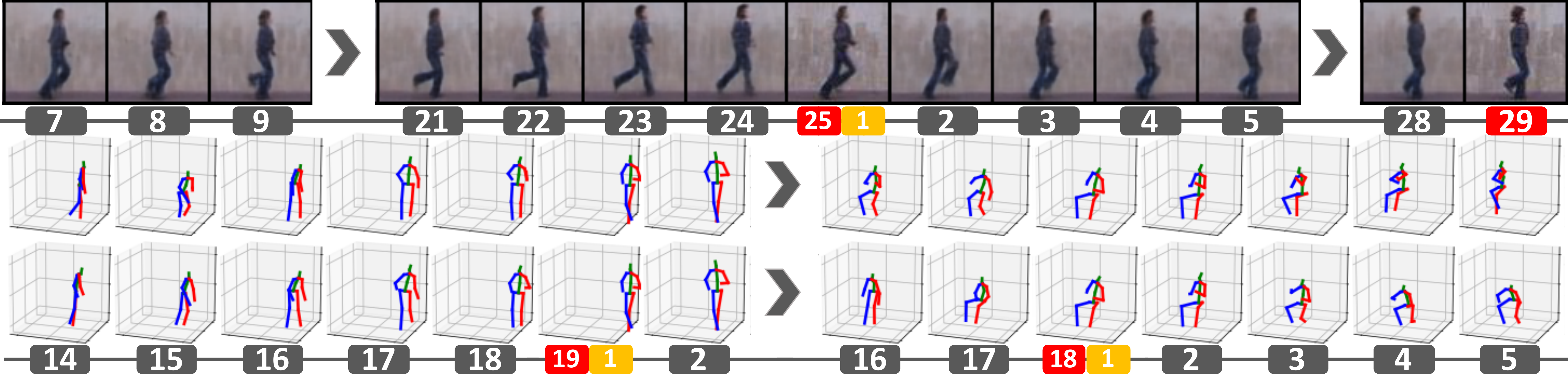}
    \vspace{-1.5em}
    \caption{Given multiple pairs of {(orange) start-} and {(red) end-}frames, we can merge multiple generated clips into a longer video, which is similar to the modern video editing process. The number beneath each frame indicates the timestamp.}
    \label{fig:bp_gen}
    \vspace{-1.5em}
\end{figure*}

\begin{figure}[t]
    \centering
    \includegraphics[width=\linewidth]{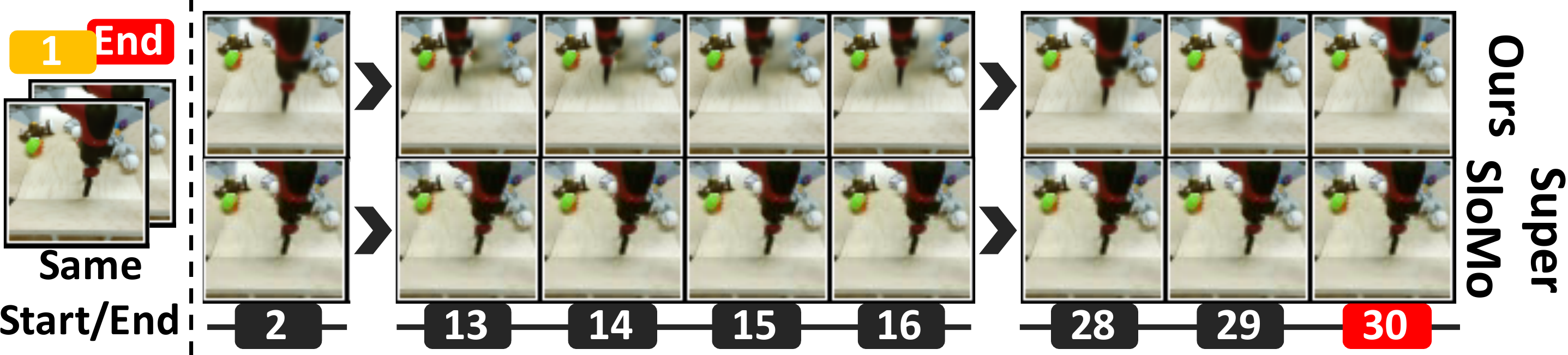}
    \vspace{-1.5em}
    \small{
        \caption{Automatic looping generation on BAIR Pushing.}
        \label{fig:loop_vidinter_bair}
        \vspace{-1.0em}
    }
\end{figure}

\begin{figure}[t]
    \centering
    \captionsetup{font=small}
    \includegraphics[width=\linewidth]{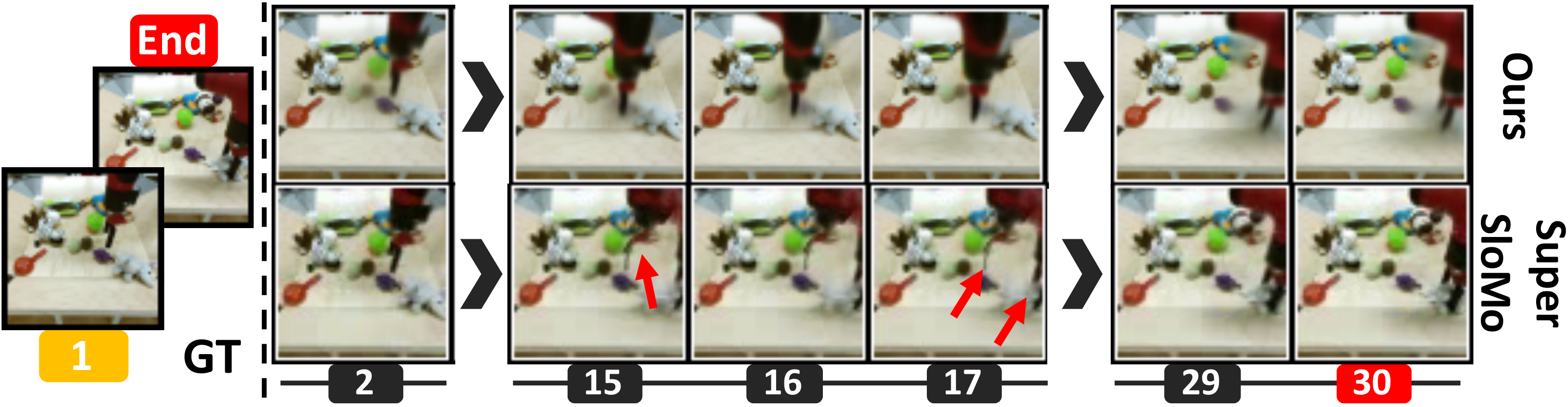}
    \vspace{-1.5em}
    \caption{Longer time-interval video generation on BAIR.} 
    \label{fig:longinteral_vidinter_bair}
    \vspace{-1.5em}
\end{figure}



To evaluate the effectiveness of our method, we conduct qualitative and quantitative analysis on \charles{four} datasets: SM-MNIST~\cite{denton2018stochastic}, Weizmann Action~\cite{ActionsAsSpaceTimeShapes_pami07}, Human3.6M~\cite{h36m_pami} and BAIR Robut Pushing~\cite{ebert17sna} to measure the CPC, quality and diversity. The following section is organized as follows: we start by stating the datasets in \secref{ssec:datasets} and evaluation metrics in \secref{ssec:eval-metrics}; the quantitative results are shown in \ref{ssec:quantitative_results}-\ref{ssec:cpc_on_post}; the qualitative results are presented in \secref{ssec:qualitative_results}; finally \charles{the comparison with VI are discussed in \ref{ssec:vi_comparison}} 


\subsection{Datasets}
\vspace{-0.4em}
\label{ssec:datasets}
We evaluate our methods on \charles{four} common testbeds: \textbf{Stochastic Moving MNIST} is introduced by \cite{denton2018stochastic} (a modified version from \cite{srivastava2015unsupervised}). The training sequence is generated by sampling one or two digits from the training set of MNIST, and then the trajectory is formed by sampling starting locations within the frame and an initial velocity vector, $(\Delta x, \Delta y) \in [-4, 4] \times [-4, 4]$. Velocity vector will be re-sampled each time the digits reach the border. \textbf{Weizmann Action} contains 90 videos of 9 people performing 10 actions. We center-crop each frame and follow the setting in \cite{he2018probabilistic} to form the training and test sets. \textbf{Human3.6M} is a large-scale dataset with 3.6 million 3D human poses captured by 11 professional actors, providing more than 800 sequences in total. We use normalized 3D skeletons of 17 joints for experiments. Following \cite{yan2018mt}, we use subjects number 1, 5, 6, 7, and 8 for training and subjects 9 and 11 for testing. \charles{\textbf{BAIR Robut Pushing~\cite{ebert17sna}} contains a robotic arm moving randomly to push diverse objects. With the large degree of stochasticity and cluttered background, it is widely used for evaluation.}

\subsection{Evaluation Metrics}
\vspace{-0.4em}
\label{ssec:eval-metrics}
We measure the structural similarity (SSIM) and peak signal-to-noise aatio (PSNR) for SM-MNIST, Weizmann \charles{and BAIR} following \cite{babaeizadeh2018stochastic,denton2018stochastic,finn2016unsupervised}. For Human3.6M, we calculate mean squared error (MSE) following \cite{yan2018mt}. To assess the learning of $p_\psi$ and $q_\phi$, we adopt the concept of \cite{yan2018mt} by introducing \textit{Sampling} and \textit{Reconstruction} metric (referred to as ``S-'' and ``R-''), where the evaluation is performed on generation from prior and posterior respectively. For each test sequence, we generate 100 samples and compute the following metrics in $95\%$ confidence interval:
\vspace{-0.3em}
\begin{itemize}[leftmargin=*]
  \setlength\itemsep{0.1em}
  \item\textbf{Control Point Consistency (S-CPC)}: We compute the mean SSIM/PSNR/MSE between the generated end-frame and the targeted end-frame since CPC should be achieved for all samples.
  \item\textbf{ Quality (S-Best)}: The best SSIM/PSNR/MSE among all samples following \cite{babaeizadeh2018stochastic,denton2018stochastic,yan2018mt}. It is a better way to assess the quality for stochastic methods because the best sample's score allows us to check if the true outcome is included in the generations.
  \item\textbf{ Diversity (S-Div)}: Adapting the concept from \cite{yang2019diversity}, we compute the variance of SSIM/PSNR across all samples with the ground-truth sequences as reference. For MSE, we calculate the variance of difference between generated and ground-truth sequences instead because MSE only measures the distance between joints while ignoring their relative positions, which will result in biased estimation for diversity.
\end{itemize}
\vspace{-1.0em}


\subsection{Quantitative Results}
\vspace{-0.4em}
\label{ssec:quantitative_results}
We show quantitative analysis on generation quality, diversity, and CPC over SM-MNIST, Weizmann, Human3.6M and \charles{BAIR -- in Table~\ref{tab:quantitative_mnist},~\ref{tab:quantitative_weizmann},~\ref{tab:quantitative_h36m}, as well as the comparison with more baselines in \tabref{tab:quantitative_bair}}. From R-Best we know that the posteriors learn well in all setting. In Table~\ref{tab:quantitative_mnist},~\ref{tab:quantitative_weizmann},~\ref{tab:quantitative_h36m}, the model with CPC+Alignment losses (\textit{+C+A}) outperforms model with only CPC loss (\textit{+C}) in S-CPC. This shows the effectiveness of alignment loss. Recall from \secref{ssec:global_and_counter} that there are two LSTMs that separate the encoder and decoder, the alignment loss aligns the two latent spaces to alleviate the mismatch between the encoding and the decoding process. Moreover, the model (\textit{Ours}) with skip-frame training further improves over \textit{+C+A} in S-CPC, where the gain mainly results from a better usage of time counter. Finally, S-CPC gain in Weizmann is less than SM-MNIST and Human3.6M since unlike the latter two, its data are captured in non-clear background with visible noise that is more challenging for CPC. \charles{On the other hands, when compared with more baselines~\cite{babaeizadeh2018stochastic,denton2018stochastic}, our method successfully models the robot's movements while maintaining CPC without hurting diversity and quality as shown in \tabref{tab:quantitative_bair}.}


For generation quality, all \charles{four} tables show comparable results in S-Best, showing that our method is able to maintain the quality while achieving CPC. Besides, the S-Best in \tabref{tab:quantitative_h36m} demonstrates an interesting finding that \textit{Ours} not only achieves extremely superior performance in S-CPC but also in S-Best. The main reason is that Human3.6M contains 3D skeletons with highly diverse actions, giving rise to considerably flexible generation. A long-term generation may easily deviate from the others, causing high S-Best error, but our method gradually converges to the targeted end-frames, confining the S-Best error (see \secref{ssec:div_through_time}). 

For generation diversity, our method attains either comparable or better performance in \tabref{tab:quantitative_mnist} and ~\ref{tab:quantitative_h36m}. This proves that our method generate diverse samples while reaching the same targeted end-frame. However, our method suffers from a larger performance drop on S-Div in \tabref{tab:quantitative_weizmann}. This is expected since Weizmann data often involve video sequences with unvarying actions, \eg walking in a fixed speed, and therefore, posing constraints at the end-frame significantly reduces the possible generation trajectories, thus leading to low diversity. Overall, our method has a significant improvement on CPC while reaching comparable generation quality and diversity with the baseline.
\vspace{-0.6em}


\subsection{CPC in Generation with Various Lengths}
\vspace{-0.4em}
\label{ssec:cpc_diff_len}

    We show the CPC performance of all models under generation of different lengths on Human3.6M dataset in \figref{fig:dynlen}. The models achieve CPC under various lengths even though they have only seen the sequences with length around $30$, showing that our models generalize well to various lengths. It is worth noting that with skip-frame training (red line), our model achieves CPC even further compared with other variations since it is able to leverage the information provided from the time counter. However, our method performs a bit worse at length 10 comparing to longer lengths because the model has less time budget for planning its trajectory and the training data do not contain any sequences with length less than 20.
    \vspace{-0.6em}


\subsection{Diversity Through Time}
\vspace{-0.4em}
\label{ssec:div_through_time}
    We evaluate the diversity of our method by investigating its behaviour through time in \figref{fig:diversity}. The downward trend can be observed around the end of the green line, which means it tries to reach the targeted end-frame as the time-counter approaches the end. However, with the skip-frame training (red line), the diversity becomes higher around the middle segment and converges near the start- and end-frame. Our full model knows its precise status such as how far it is to the end-frame or how much time budget remains, and thus can plan ahead to achieve CPC. Since our model perceives well about its time budget, it can ``go wild'', \ie, explore all possible trajectories while still being capable of getting back to the targeted end-frame on time. 
    \vspace{-0.6em}


\subsection{CPC Weight on Prior vs. Posterior}
\vspace{-0.4em}
\label{ssec:cpc_on_post}
   We assess the effect of posing different CPC weights on prior $p_{\psi}$ versus posterior $q_{\phi}$ by comparing the quality and diversity in SSIM (\figref{fig:cpc_prior_good}). With different weights, the behavior of diversity for $p_{\psi}$ and $q_{\phi}$ is comparable. However, CPC on $p_{\psi}$ (blue line) does not result in degradation throughout all CPC weights in comparison with posing CPC on $q_{\phi}$. This shows that our method is more robust to different CPC weights.  
   \vspace{-0.6em}


\subsection{Qualitative Results}
\vspace{-0.4em}

\label{ssec:qualitative_results}
\paragraph{Generation with various lengths.} In Fig.~\ref{fig:dynlen_bair}, we show roughly how p2p generation works by comparing with \cite{denton2018stochastic} on BAIR dataset.  Fig.~\ref{fig:vary_len} shows various examples across other datasets. Our model maintains high CPC for all lengths while producing diverse results.
\vspace{-1.2em}

\paragraph{Multiple control-points generation.}
In \figref{fig:bp_gen}, we show the generated videos with multiple control points. The first row highlights transition across different attributes or actions (\ie, ``run'' to ``skip'' in Weizmann dataset). The second and third rows show two generated videos with the same set of multiple control points (\ie, stand; sit and lean to the left side). Note that these are two unique videos with diverse frames in transitional timestamps. By placing each control point as a breakpoint in a generation, we can achieve fine-grained controllability directly from frame exemplars. 
\vspace{-1.2em}

\paragraph{Loop generation.} \figsref{fig:loop_gen} \& \ref{fig:loop_vidinter_bair} show that our method can be used to generate infinite looping videos by forcing the targeted start- and end-frames to be the same.
\vspace{-0.3em}

\footnotetext[1]{We thank MOST-107-2634-F-007-007, MOST-106-2221-E-007-080-MY3, MOST Joint Research Center for AI Technology, and All Vista Healthcare for their supports.}

\subsection{Comparison with Video Interpolation}
\vspace{-0.4em}
\label{ssec:vi_comparison}
To elaborate the essential difference between VI and p2p generation, we conduct a task of inserting 28 frames between start- and end-frame where the temporal distance between the targeted start- and end-frames is large (\figref{fig:longinteral_vidinter_bair}). Note that Super SloMo~\cite{jiang2018super} produces artifacts such as distortion or two robot arms (indicated by red arrows in the 15th and 17th frames). VI methods typically are deterministic approaches while p2p generation is able to synthesize diverse frames (see \figref{fig:dynlen}). Finally, automatic looping can be accomplished by p2p generation but not by VI. Given the same start- and end-frames, we confirm that Super SloMo~\cite{jiang2018super} will interpolate all the same frames as if the video is freezing (\figref{fig:loop_vidinter_bair}).
\vspace{-0.3em}

\section{Conclusion}
\vspace{-0.6em} 
\comment{
\todo{
\begin{itemize}
    \item We propose p2p 
    \item We propose a method to solve p2p
    \item In a nutshell, we have a model which is capable of achieving CPC under various length without loss of quality and diversity.
    \item Future work
\end{itemize}
}
}

The proposed point-to-point (p2p) generation controls the generation process with two control points---the targeted start- and end-frames---to provide better controllability in video generation. To achieve control point consistency (CPC) while maintaining generation quality and diversity, we propose to maximize the modified variational lower bound for conditional video generation model, followed by a novel skip-frame training strategy and a latent space alignment loss to further reinforce CPC. We show the effectiveness of our model via extensive quantitative analysis. The qualitative results further highlight the merits of p2p generation. However, our current model cannot handle high-resolution videos. Modeling all the details such as small objects or noisy background in high-res videos is still an open problem for the existing video generation/prediction methods. We will explore this direction in the future.
Overall, our work opens up a new dimension in video generation that is promising for further exploration.


{\small
\bibliographystyle{ieee_fullname}
\bibliography{citation}
}

\clearpage

\begin{appendix}


\section{Overview}
\todo{
The supplementary material is organized as follows:

\noindent First, we provide an overview video (video link: \url{https://drive.google.com/open?id=1kS9f2oNGFPO_hp7iWZmvtXLPnOrhl9qW}), which briefly summarizes our work. Second, we provide more quantitative results on all datasets: SM-MNIST, Weizmann Human Action, and Human3.6M in \secref{sec:quan_res}. Furthermore in \secref{sec:qual_res}, we present more qualitative evaluations with respect to {\em i}) ``Generation with various length'' in Figs.~\ref{fig:supp_dynlen_smmnist}-\ref{fig:supp_dynlen_h36m} (more examples at \url{https://drive.google.com/open?id=1ueQHNx56MWoqL9ilHjZuBZourg4VrbKc}); {\em ii}) ``Multiple control-points generation'' in \figref{fig:supp_bpgen} (more examples at \url{https://drive.google.com/open?id=1OUOd2LjmKwHwVpRwldUEIgvzfpWucYjt}); {\em iii}) ``Loop generation'' in \figref{fig:supp_loopgen} (more examples at \url{https://drive.google.com/open?id=1kb8PCIR2_lkE1JS6NlwyglxKlChSBSbF}). Finally, the implementation details are described in \secref{sec:implementation_details}.
}

\section{Quantitative Results}
\label{sec:quan_res}
\subsection{Performance Under Various Length}

\begin{figure*}[t!]
    \centering
    \includegraphics[width=\linewidth]{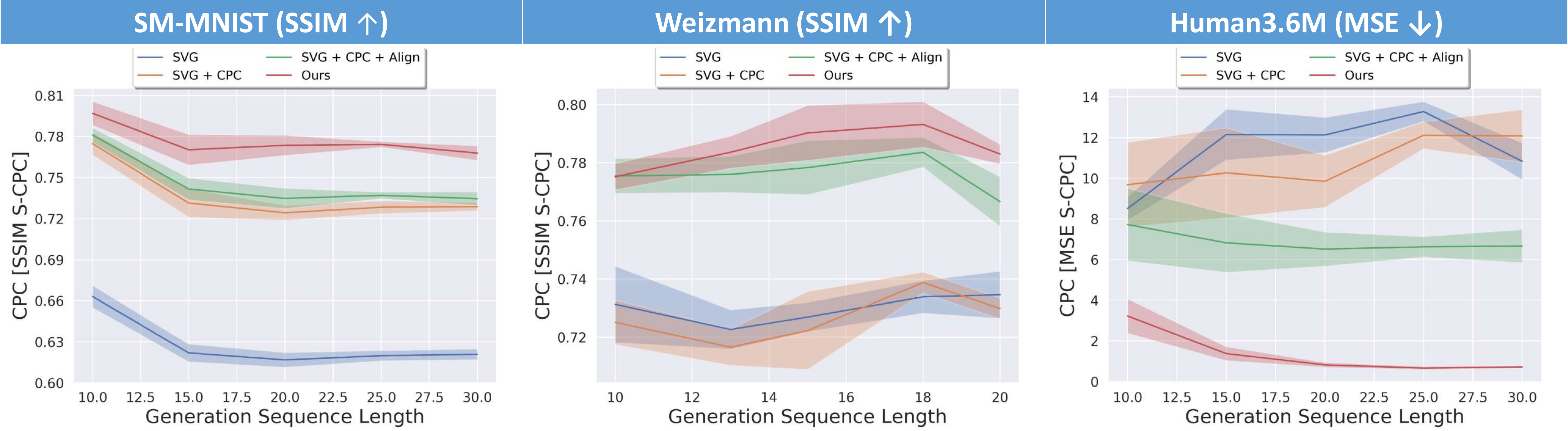}\vspace{-0.5em}
    \caption{(Better view in color) Control point consistency in generation with various length. \todo{Our model significantly outperforms other baselines on all datasets under various lengths.}}
    \label{fig:supp_dynlen_cpc}
    \vspace{-0.5em}
\end{figure*}

\begin{figure*}[t!]
    \centering
    \includegraphics[width=\linewidth]{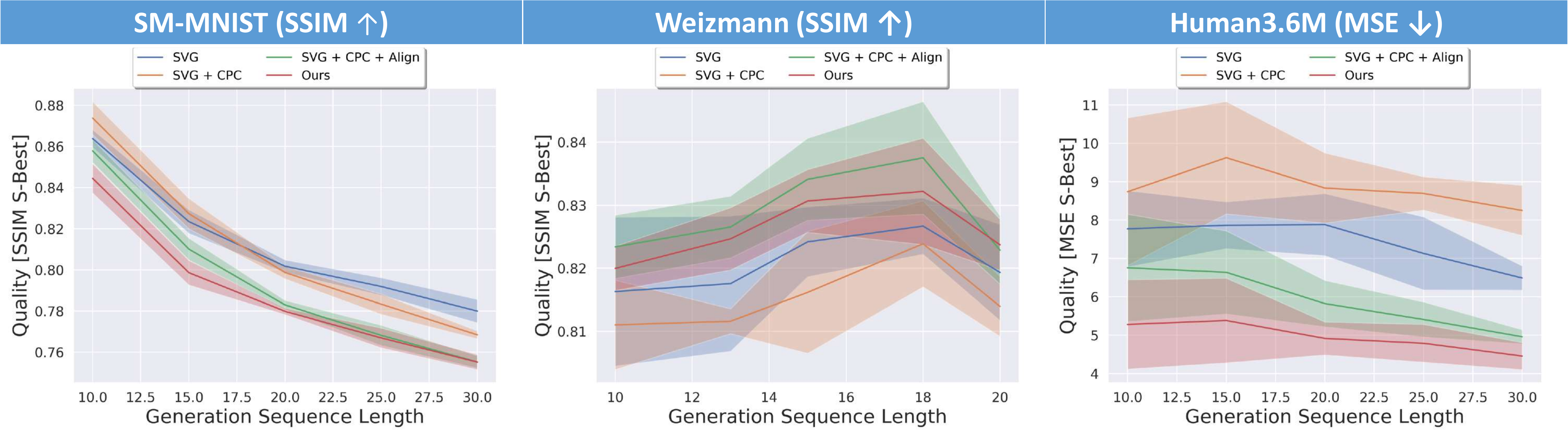}\vspace{-0.5em}
    \caption{(Better view in color) Quality in generation with various length. \todo{Our model sustains the generation quality on the three datasets while achieving CPC.}}
    \label{fig:supp_dynlen_quality}
    \vspace{-0.5em}
\end{figure*}

\begin{figure*}[t!]
    \centering
    \includegraphics[width=\linewidth]{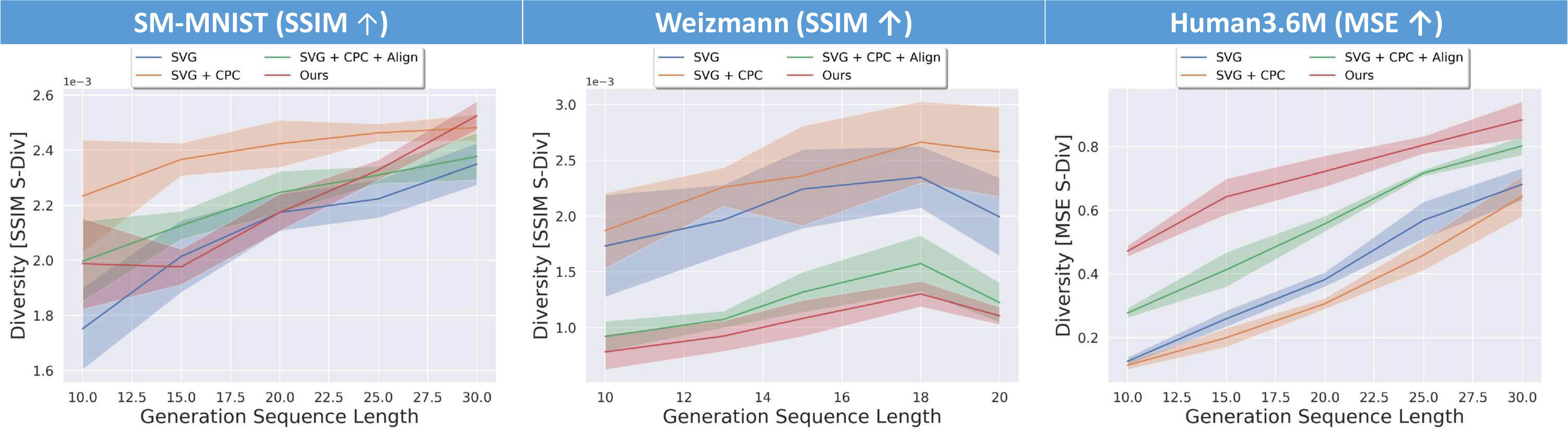}\vspace{-0.5em}
    \caption{(Better view in color) Diversity in generation with various length. \todo{Ours achieve better or comparable diversity on SM-MNIST and Human3.6M while achieving CPC.}}
    \label{fig:supp_dynlen_diversity}
    \vspace{-0.5em}
\end{figure*}

\begin{figure*}[t!]
    \centering
    \includegraphics[width=\linewidth]{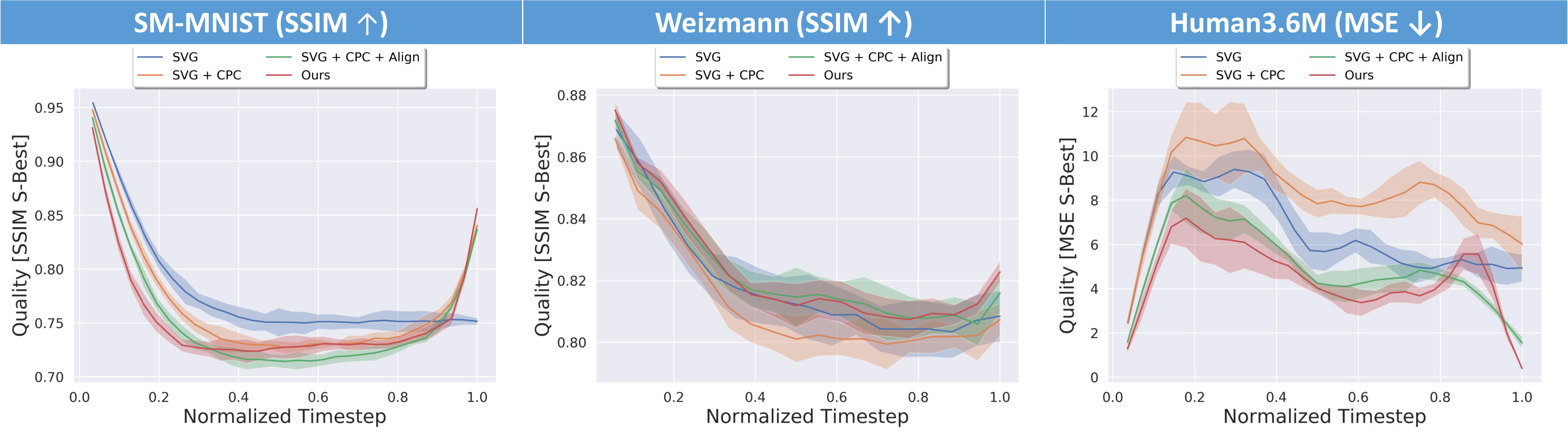}\vspace{-0.5em}
    \caption{(Better view in color) Quality through time. \todo{The generation quality of our model is comparable on SM-MNIST, Weizmann and better on Human3.6M while achiving control-point consistency.}}
    \label{fig:supp_timestep_quality}
    \vspace{-0.5em}
\end{figure*}

\begin{figure*}[t!]
    \centering
    \includegraphics[width=\linewidth]{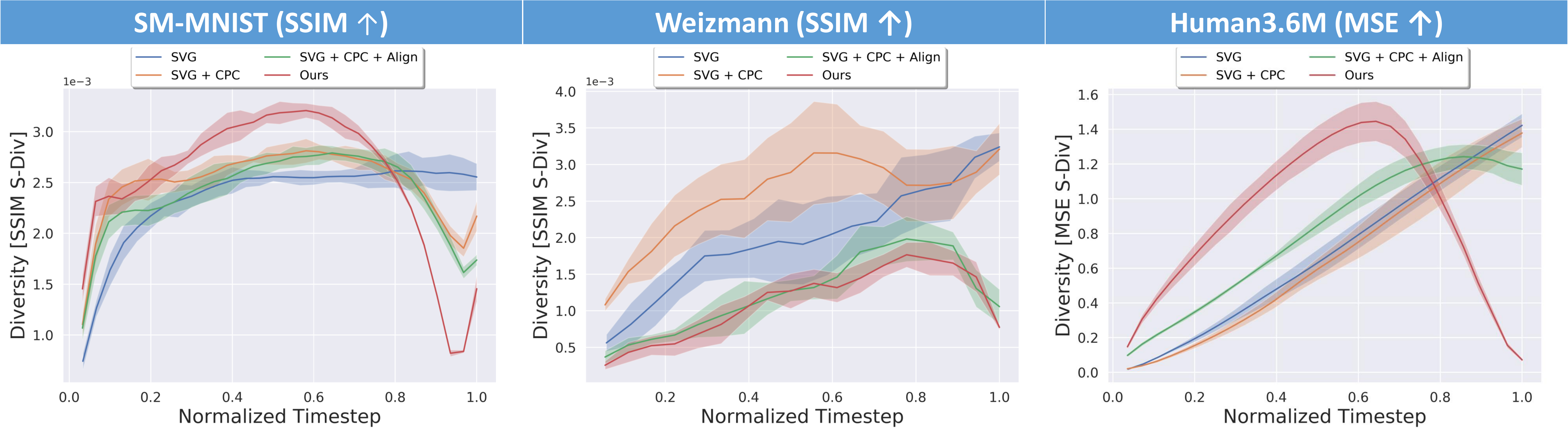}\vspace{-0.5em}
    \caption{(Better view in color) Diversity through time. \todo{The diversity is high in the intermediate frames but reaches zero at the two control points---the targeted start- and end-frames.}}
    \label{fig:supp_timestep_diversity}
    \vspace{-0.5em}
\end{figure*}

In this section, we investigate control point consistency, generation quality and diversity under generation of different lengths on SM-MNIST, Weizmann Action, and Human3.6M dataset (refer to Sec. 4.4 in the main paper).

\paragraph{Control Point Consistency (S-CPC):} In \figref{fig:supp_dynlen_cpc}, we show the performance of CPC on the three datasets, where for SM-MNIST and Weizmann (the first and the second column), the higher (SSIM) the better, and for Human3.6M (the last column), the lower (MSE) the better. Our method (red line) significantly outperforms other baselines on all datasets, while different components of our method including CPC on prior, latent space alignment, and skip-frame training all introduce performance gain. 

\paragraph{Quality (S-Best):} In \figref{fig:supp_dynlen_quality}, we demonstrate that our method is able to sustain the generation quality on the three datasets, with the higher (SSIM) the better for SM-MNIST and Weizmann (the first and the second column), and the lower (MSE) the better for Human3.6M (the last column). Our method (red line) achieves superior quality on Human3.6M since its data contain 3D skeletons with highly diverse actions and imposing a targeted end-frame largely confines the S-Best error (more details mentioned in Sec. 4.5 in the main paper). On the other hand, for SM-MNIST and Weizmann, our method only suffers from marginal performance drop in comparison with other baselines. We point out that the generation quality in SM-MNIST gradually declines with increasing generation length since the two digits are prone to overlapping with each other in a longer sequence, resulting in blurry generation after the encounter. This can be potentially solved by representation disentanglement \cite{villegas2017decomposing,denton2017unsupervised,tulyakov2018mocogan,hsieh2018learning,wiles2018self}, which is out of scope of this paper and left to future work. Overall, we establish that our method attains comparable generation quality while achieving CPC.

\paragraph{Diversity (S-Div):} Finally, we show the generation diversity on the three datasets in \figref{fig:supp_dynlen_diversity}, where for all columns, the higher (SSIM or MSE) the better. We can observe that our method (red line) reaches superb and comparable performance on Human3.6M and SM-MNIST dataset respectively. On the contrary, Weizmann dataset involves video sequences with steady and fixed-speed action and hence tremendously reduces the possibility of generation if posing constraint at the end-frame (red line in the middle column). All in all, regardless of the limitation of dataset itself, our method is capable of generating diverse sequences and simultaneously achieving CPC.

\subsection{Performance Through Time}

In this section, we perform a more detailed analysis on generation quality and diversity through time (refer to Sec. 4.5 in the main paper).

\paragraph{Quality (S-Best):} In \figref{fig:supp_timestep_quality}, we show the generation quality at each timestep on the three datasets, with the higher (SSIM) the better for SM-MNIST and Weizmann (the first and second columns), and the lower (MSE) the better for Human3.6M (the last column). We can observe a consistent trend across methods and datasets that the quality progressively decreases as the timestep grows. This is expectable since the generated sequences will step-by-step deviate from the ground truth and induce compounding error as the generation is gradually further from the given start-frame. Remarkably, for all methods taking CPC into consideration (orange, green, and red lines), there is a strong comeback on the generation quality at the end of the sequence since achieving CPC ensures that the generated end-frame converges to the targeted end-frame, thus leading to the results with better S-Best at the last timestep. Finally, the quality boost at the end-frame is lower in Weizmann dataset (the middle column) since unlike the other two (the first and the last columns), its data are captured in noisy background, posing more challenges to CPC and consequently causing lower quality at the end frame.

\paragraph{Diversity (S-Div):} In \figref{fig:supp_timestep_diversity}, we demonstrate the generation diversity through time on the three datasets, with the higher (SSIM or MSE) the better in all columns. A consistent trend is shared across all datasets (all columns) in our method (red line) where the diversity is high in the intermediate frames but reaches zero at the two control points---the targeted start- and end-frames. This suggests that our method is able to plan ahead, generate high-diversity frames at the timestep far from the end, and finally converge to the targeted end-frame with zero-approaching diversity. In addition, we point out that the diversity curve of Weizmann dataset (the middle column) indicates a slightly worse performance in comparison to the results on the other two datasets (the first and third columns) since Weizmann data is featured by unvarying actions, \eg, walking in a fixed speed, that immensely reduces the potential diversity at the intermediate frames.

\section{Qualitative Results}{
\label{sec:qual_res}
\begin{figure*}[t!]
    \centering
    \includegraphics[width=\linewidth]{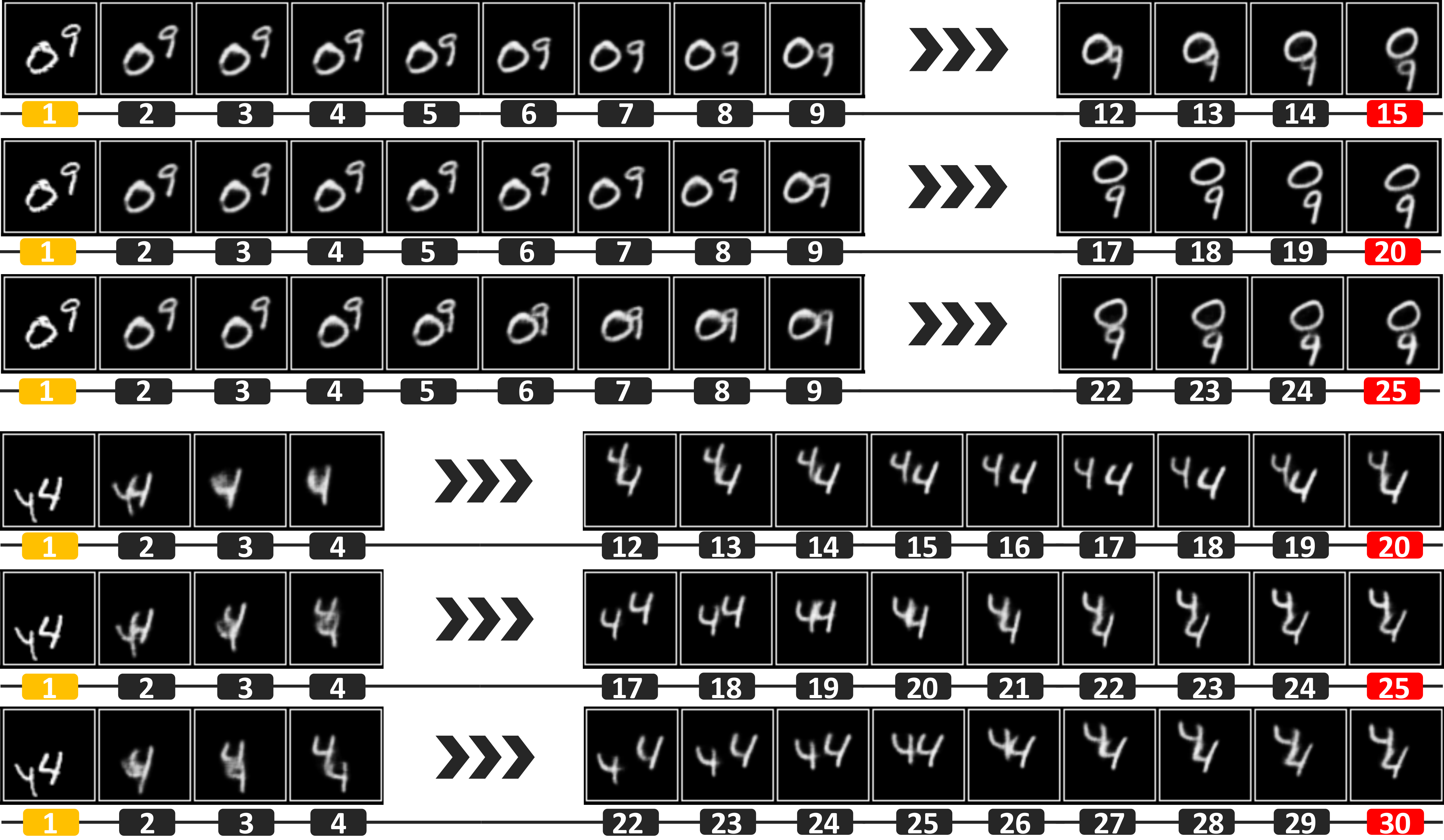}\vspace{-0.5em}
    \caption{Generation with various length on SM-MNIST. \todo{Given the targeted (orange) start- and (red) end-frames, we show the generation results with various lengths on SM-MNIST.}}
    \label{fig:supp_dynlen_smmnist}
    \vspace{-1.0em}
\end{figure*}

\begin{figure*}[t!]
    \centering
    \includegraphics[width=\linewidth]{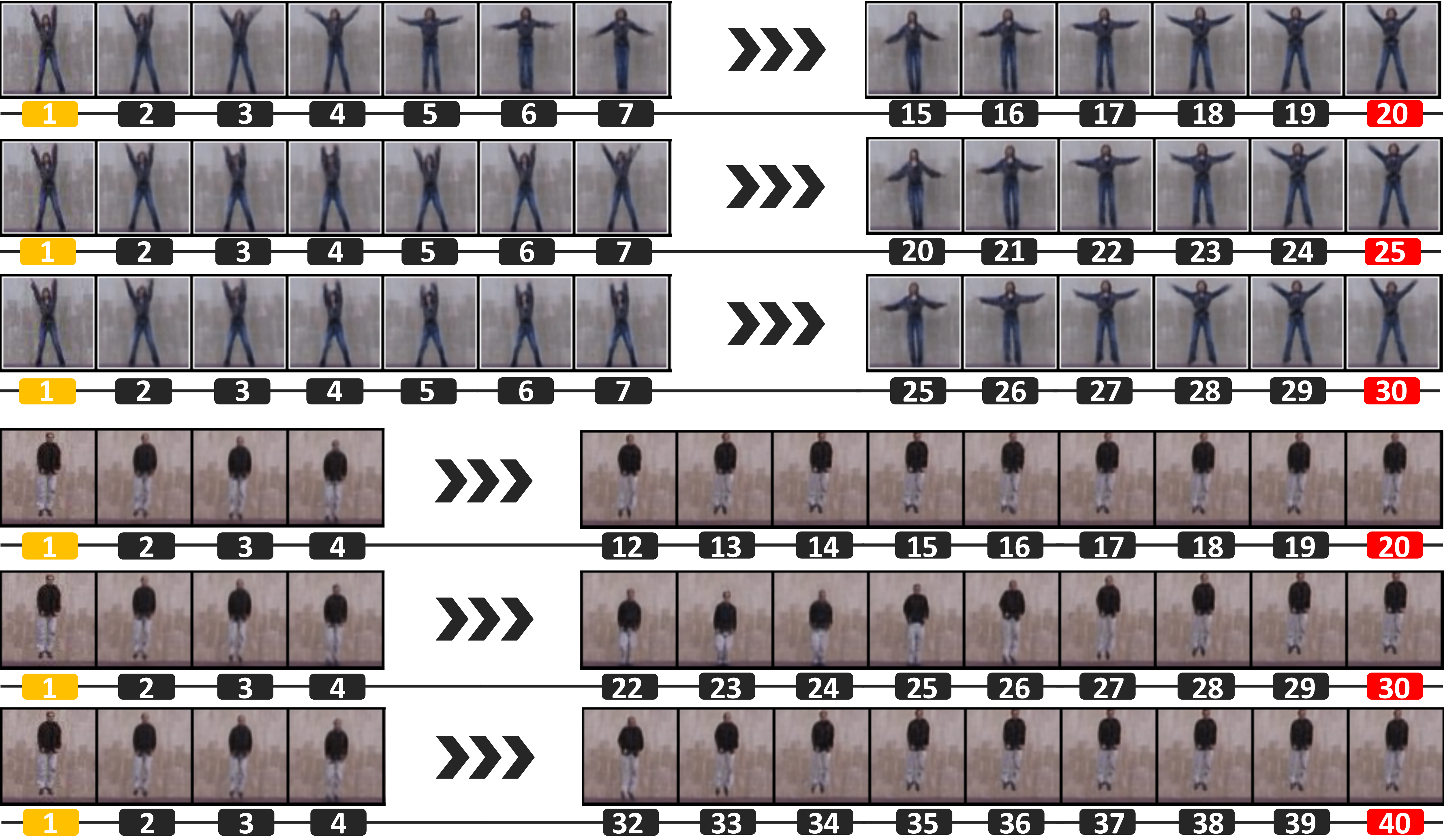}\vspace{-0.5em}
    \caption{Generation with various length on Weizmann. \todo{Given the targeted (orange) start- and (red) end-frames, we show the generation results with various lengths on Weizmann.}}
    \label{fig:supp_dynlen_weizmann}
    \vspace{-1.0em}
\end{figure*}

\begin{figure*}[t!]
    \centering
    \includegraphics[width=\linewidth]{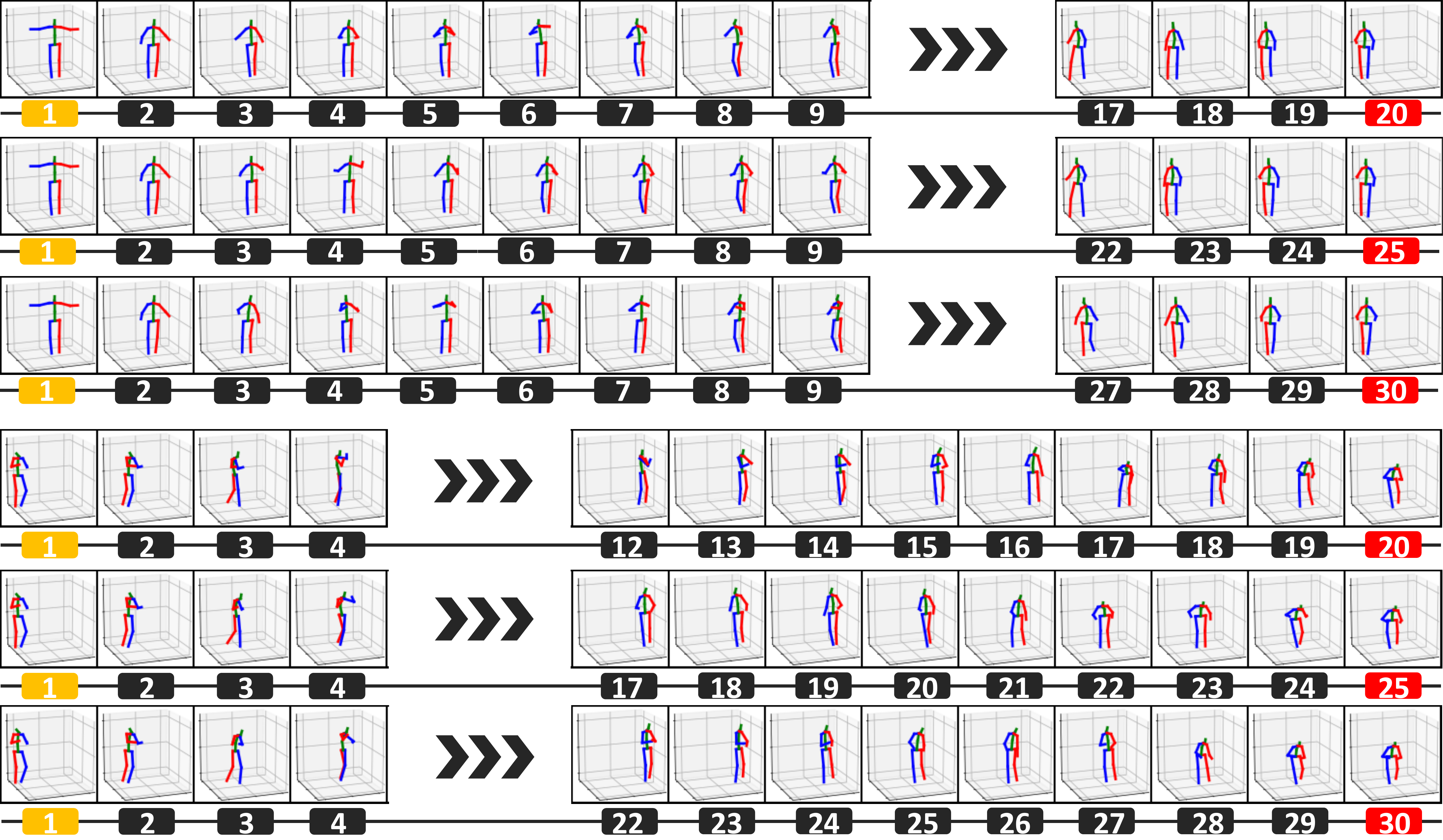}\vspace{-0.5em}
    \caption{Generation with various length on Human3.6M. \todo{Given the targeted (orange) start- and (red) end-frames, we show the generation results with various lengths on Human3.6M.}}
    \label{fig:supp_dynlen_h36m}
    \vspace{-1.0em}
\end{figure*}

\begin{figure*}[t!]
    \centering
    \includegraphics[width=\linewidth]{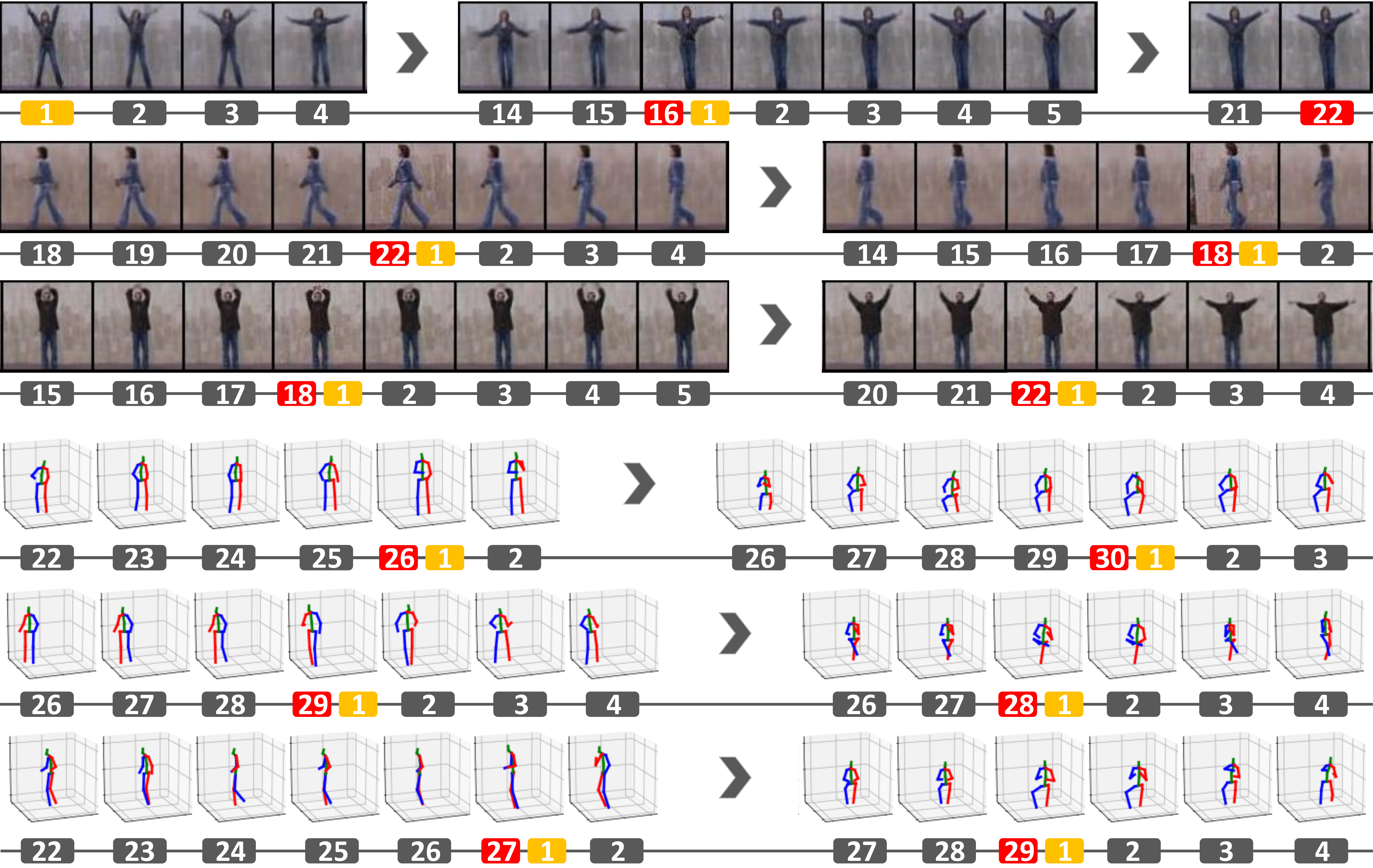}\vspace{-0.5em}
    \caption{Multiple control points generation. \todo{Given multiple targeted (orange) start- and (red) end-frames, we can merge multiple generated clips into a longer video.}}
    \label{fig:supp_bpgen}
    \vspace{-0.5em}
\end{figure*}

\begin{figure*}[t!]
    \centering
    \includegraphics[width=\linewidth]{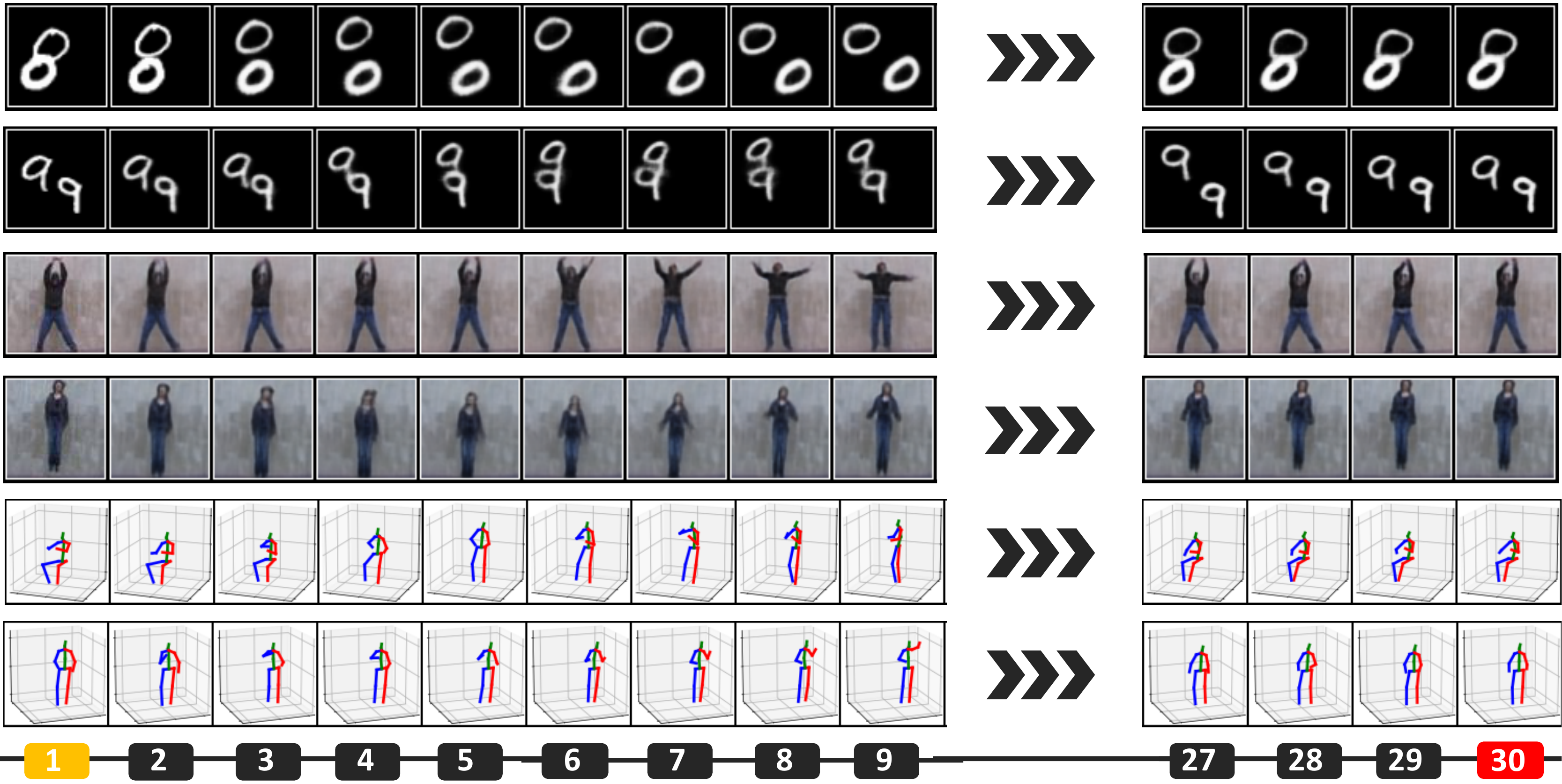}\vspace{-0.5em}
    \caption{Loop generation. \todo{We set the targeted (orange) start- and end-frame with the same frame to achieve loop generation.}}
    \label{fig:supp_loopgen}
    \vspace{-0.5em}
\end{figure*}

\paragraph{Generation with various length.} \todo{In \figref{fig:supp_dynlen_smmnist}, \figref{fig:supp_dynlen_weizmann}, and \figref{fig:supp_dynlen_h36m}, we demonstrate the generation results with various lengths on SM-MNIST, Weizmann Action, and Human3.6M datasets. For more generated examples, please see \url{https://drive.google.com/open?id=1ueQHNx56MWoqL9ilHjZuBZourg4VrbKc}.}

\paragraph{Multiple control-points generation.} \todo{In \figref{fig:supp_bpgen}, given multiple targeted start- and end-frames, we show our model's ability to merge multiple generated clips into a longer video. For more generated examples, please see \url{https://drive.google.com/open?id=1OUOd2LjmKwHwVpRwldUEIgvzfpWucYjt}.}

\paragraph{Loop generation.} \todo{In \figref{fig:supp_loopgen}, by setting the targeted start- and end-frame to be the same, we can achieve loop generation. For more generated examples, please see \url{https://drive.google.com/open?id=1kb8PCIR2_lkE1JS6NlwyglxKlChSBSbF}.}
}

\section{Implementation Details}
\label{sec:implementation_details}
We provide the training details and network architecture in this section.

\subsection{Training Details}
We implement our model in PyTorch. For SM-MNIST and Weizmann Action the input and output image size is $64 \times 64$, and for Human3.6M the input comprises the joint positions of size $17 \times 3$. Note that while our p2p generation models are fed with the targeted end-frames, the baseline method SVG~\cite{denton2018stochastic}, which is not CPC-aware, is introduced with one additional frame such that all methods are compared under the same number of input frames.
For the reconstruction loss in $\mathcal{L}_{\theta, \phi, \psi}^{\mathrm{full}}$, we use $L_2$-loss. All models are trained with Adam optimizer, learning rate of $0.002$, and batch size of $100, 64, 128$ for SM-MNIST, Weizmann Action and Human3.6M respectively. The weights in the full objective function and other details regarding each dataset are summarized as follows:

\paragraph{SM-MNIST:}{
For the weights in $\mathcal{L}_{\theta, \phi, \psi}^{\mathrm{full}}$, we set $\beta=10^{-4}$, $\alpha_\mathrm{cpc}=100$, $\alpha_\mathrm{align}=0.5$, $p_{skip}=0.5$.
And the length of training sequences is $12\rpm3$.

\paragraph{Weizmann Action:}{
For the weights in $\mathcal{L}_{\theta, \phi, \psi}^{\mathrm{full}}$, we set $\beta=10^{-5}$, $\alpha_\mathrm{cpc}=10^5$, $\alpha_\mathrm{align}=0.1$, $p_\mathrm{skip}=0.3$. The length of training sequences is $15\rpm3$ for Weizmann Action and we augment the dataset by flipping each sequence so that our model can learn to generate action sequences that proceed toward both directions.
}

\paragraph{Human3.6M:}{
For the weights in the objective function $\mathcal{L}_{\theta, \phi, \psi}^{\mathrm{full}}$: $\beta=10^{-5}$, $\alpha_\mathrm{cpc}=10^5$, $\alpha_\mathrm{align}=1.0$, $p_\mathrm{skip}=0.3$.
The length of training sequences for Human3.6M is $27\rpm3$. Besides, we speed up the training sequences to $6\times$ since the adjacent frames in the original sequences are often too similar to each other, which may prevent the model from learning diverse actions.
}

}

\subsection{Network Architecture}
The networks for three datasets all contain the following main components: {\em i}) posterior $q_\phi$, {\em ii}) prior $p_\psi$, and {\em iii}) generator $p_\theta$. The encoder is shared by $q_\phi$, $p_\psi$ and the global descriptor. We choose DCGAN \cite{radford2016unsupervised} as the backbone of our encoder and decoder for SM-MNIST and Weizmann Action, and choose multilayer perceptron (MLP) for Human3.6M. The hyper-parameters for the decoder, encoder, $q_\phi$, $p_\psi$ and $p_\theta$ for each dataset are listed below:
    \paragraph{SM-MNIST:}
        For the networks we set $|h_t|=128$, $|z_t|=10$; one-layer, $256$ hidden units for $q_\phi$, one-layer, $256$ hidden units for $p_\phi$, two-layer, $256$ hidden units for $p_\theta$.
    \paragraph{Weizmann Action:}
        We use $|h_t|=512$, $|z_t|=64$; one-layer, $1024$ hidden units for $q_\phi$, one-layer, $1024$ hidden units for $p_\phi$, two-layer, $1024$ hidden units for $p_\theta$.
    \paragraph{Human3.6M:}
        The networks have $|h_t|=512$, $|z_t|=32$; one-layer, $1024$ hidden units for $q_\phi$, one-layer, $1024$ hidden units for $p_\phi$, two-layer, $1024$ hidden units for $p_\theta$. The encoder MLP consists of 2 residual layers with hidden size of $512$, followed by one fully-connected layer and activated by $\mathrm{tanh}$ function; the decoder MLP is the mirrored version of the encoder but without $\mathrm{tanh}$ in the output layer.
    
\end{appendix}

\end{document}